\documentclass{article}
\usepackage[T1]{fontenc}
\usepackage{calr_preprint,times}
\usepackage{amsmath,amssymb}
\usepackage{booktabs}
\usepackage{multirow}
\usepackage{xcolor}
\usepackage{array,colortbl}
\definecolor{MRgroup}{HTML}{FCEBD5}
\definecolor{MRunified}{HTML}{E5EFFA}
\definecolor{MRdecoupled}{HTML}{E7F2E8}
\newsavebox{\MRtablebox}
\newcommand{\MRcell}[3]{\ensuremath{#1^{#3}_{\scriptscriptstyle\pm#2}}}
\newcommand{\MRbest}[1]{\mathbf{#1}}

\usepackage{graphicx}
\usepackage{microtype}
\usepackage{placeins}
\usepackage{wrapfig}
\usepackage{needspace}
\usepackage{hyperref}
\hypersetup{
  hidelinks,
  pdftitle={CALR: Continuous Anchored Latent Reasoning via Render-of-Thought Compression},
  pdfauthor={Zhaoyang Wei, Bowen Jiang, Yanchao Hao, Wenchao Ding, Zheng Wei, Shaocheng Wu, Zhenjun Han, Jianbin Jiao},
  pdfsubject={Preprint}
}
\usepackage{url}

\graphicspath{{figures/}}

\title{CALR: Continuous Anchored Latent Reasoning\\via Render-of-Thought Compression}
\author{%
Zhaoyang Wei\textsuperscript{1}\quad
Bowen Jiang\textsuperscript{1}\quad
Yanchao Hao\textsuperscript{2}\quad
Wenchao Ding\textsuperscript{2}\\[3pt]
Zheng Wei\textsuperscript{2}\quad
Shaocheng Wu\textsuperscript{2}\quad
Zhenjun Han\textsuperscript{1}\quad
Jianbin Jiao\textsuperscript{1}\\[6pt]
\textsuperscript{1}University of the Chinese Academy of Sciences\\
\textsuperscript{2}Tencent
}

\begin{document}
\raggedbottom
\maketitle

\begin{abstract}
Visual latent reasoning compresses rendered derivations into compact intermediate states, reducing textual reasoning overhead. Existing approaches differ in how they represent these states: continuous methods avoid vocabulary constraints, whereas discrete methods improve accuracy through quantization into a finite codebook.
Our analysis of representative continuous and discrete systems identifies two functional requirements: answers must rely on latent states, and those states must carry valid, problem-specific reasoning.
Continuous latents influence answers despite collapsed reasoning content, whereas discrete latents retain recoverable intermediate reasoning that answer prediction largely bypasses.
To address these challenges, we propose Continuous Anchored Latent Reasoning (CALR), which connects latent formation with answer use through functional anchoring.
With reference latents from information-balanced compression, CALR couples latent-mediated answer supervision with derivation-level semantic anchoring: the former routes answer supervision through intermediate states, while the latter grounds their decoded content in problem-specific derivations.
A parallel-to-autoregressive curriculum develops sequential reasoning by conditioning subsequent latent blocks on generated prefixes.
Under matched budgets, CALR gains 26 points over a comparable continuous latent reasoning method.
Further analyses show that its latents support answer prediction and carry problem-specific intermediate reasoning.
\end{abstract}
\section{Introduction}
\label{sec:introduction}

Chain-of-thought (CoT) enables language models to develop intermediate reasoning before producing an answer~\citep{wei2022cot}. Verbalizing that reasoning, however, requires a long sequence of decoding steps. Latent reasoning moves the intermediate solution process into compact hidden states, reducing the need to express every step in words~\citep{hao2025coconut,shen2025codi,tan2025colar}. Optical compression offers a concrete connection between these states and explicit derivations: rendering text as an image allows long passages to be encoded in substantially fewer visual tokens~\citep{wei2025deepseekocr}. Applied to reasoning traces, this motivates \emph{visual latent reasoning}, in which a model generates optical latent states to carry its solution process before answering.

 Previous approaches learn continuous visual latents from rendered reasoning traces, either aligning generated states with visual embeddings or regularizing latent distributions with visual-semantic features~\citep{wang2026regular,wang2026rot}. Motivated by collapse in continuous representations, subsequent work quantizes optical features into a discrete vocabulary and models the resulting latent codes autoregressively, achieving higher accuracy than earlier latent methods~\citep{chang2026dlr}. Both paradigms advance compact representations of reasoning, but retain distinct representational limitations: existing continuous latents are susceptible to feature collapse, whereas the gains of discrete methods come with a restricted range of latent representations determined by a finite vocabulary.

\begin{figure}[!t]
    \centering
    \includegraphics[width=\linewidth]{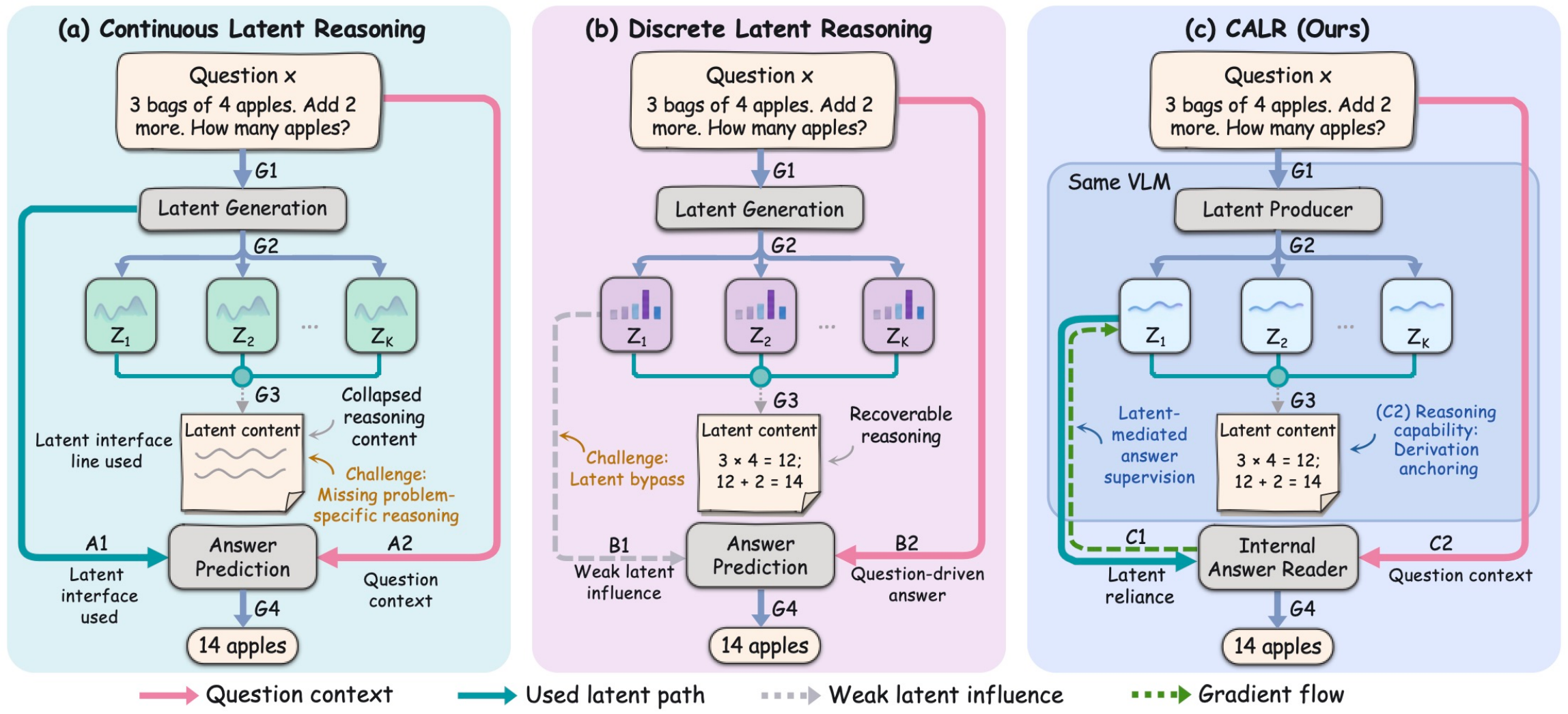}
    \vspace{-20pt}
    \caption{\textbf{The latent handoff challenge.} (a) Continuous latents can support answer prediction without carrying problem-specific reasoning. (b) Discrete latents can encode recoverable reasoning yet be bypassed by the answer model. (c) CALR couples latent reliance (C1) with derivation anchoring for reasoning capability (C2).}
    \label{fig:latent-handoff-challenge}
    \vspace{-15pt}
\end{figure}

To understand how these compact representations support problem solving, we analyze representative continuous and discrete systems through interventions and derivation decoding. Our analysis identifies two requirements for effective latent reasoning (Figure~\ref{fig:latent-handoff-challenge}): \emph{latent reliance} (C1), whereby answer prediction depends on the generated latents, and \emph{reasoning capability} (C2), whereby those latents carry a valid derivation for the current problem. In the discrete system, intermediate information can be recovered from the latents, yet answer prediction largely bypasses them. In the continuous system, the latent interface is load-bearing, yet replacing states across problems barely changes accuracy, and decoding recovers little problem-specific reasoning. 
These complementary failures reveal a \emph{missing handoff} between latent formation and answer prediction: representation supervision shapes the latent states, while question-visible answer supervision can be satisfied without using the reasoning those states are intended to carry.

To address these challenges, we propose \textbf{Continuous Anchored Latent Reasoning (CALR)}, which couples answer prediction and derivation supervision through the same continuous latent trajectory. A question-blind external reader directs answer supervision through the latents, while a frozen DeepSeek-OCR2 decoder~\citep{wei2026deepseekocr2} supervises their problem-specific reasoning content. Stage~I establishes these supervisory paths by learning an information-balanced resampler of rendered derivations and training the external reader on the resulting reference latents. Stage~II learns problem-conditioned latent generation under both frozen supervisors: parallel drafting warm-starts autoregressive refinement, in which later latent blocks build on generated prefixes. CALR supports a decoupled realization with an external answer model and a unified realization that generates latents and answers within the same VLM.

We evaluate CALR across five reasoning benchmarks and multiple model families. On GSM8K-Aug (Qwen3-VL-4B-Instruct), unified autoregressive CALR achieves 63.8\% versus RoT's 37.8\% accuracy with 32 continuous latents each, a 26.0-point gain. Interventions and decoding show that its latents support answers and carry problem-specific reasoning. Our contributions are threefold:
\begin{itemize}
    \item We establish a unified functional framework for visual latent reasoning based on latent reliance and reasoning capability, revealing complementary failures in existing approaches.
    \item We propose CALR, a unified training framework that couples latent-mediated answer supervision with derivation anchoring, jointly shaping the use and reasoning content of continuous latent trajectories.
    \item We demonstrate that continuous latents can deliver substantial accuracy gains while carrying load-bearing, problem-specific reasoning, supported by cross-model evaluations, interventions, and derivation decoding.
\end{itemize}
\section{Preliminary Analysis: The Missing Handoff in Latent Reasoning}
\label{sec:prelim-v13}

Visual latent reasoning requires a latent trajectory that the answer model uses and that carries the reasoning needed to solve the problem. We formalize these requirements as \emph{latent reliance} (C1) and \emph{reasoning capability} (C2), then examine them in continuous and discrete visual latent systems.

\subsection{Problem Formulation: Two Requirements for Visual Latent Reasoning}
\label{sec:criterion-v13}

Let $(x,y)\sim\mathcal D$ denote a question--answer pair, and let $\mathcal D_X$ be its question marginal. A visual latent reasoner consists of a latent generator $G_\theta$ and an answer model $p_\phi$,
\begin{equation}
G_\theta:\mathcal X\rightarrow\mathcal Z^K,
\qquad
p_\phi:(\mathcal X\cup\{\varnothing_x\})
\times(\mathcal Z^K\cup\{\varnothing_z\})
\rightarrow\Delta(\mathcal Y),
\label{eq:vlr-maps-v13}
\end{equation}
where $\mathcal Z$ may be a continuous visual feature space or a discrete optical code space, while $\varnothing_x$ and $\varnothing_z$ denote the absence of the question and latent inputs, respectively. The two maps specify functional roles and may share parameters. For a sequence of $K$ latent states, inference is
\begin{equation}
z_{1:K}=G_\theta(x),
\qquad
p_{\theta,\phi}(y\mid x)=p_\phi(y\mid x,z_{1:K}),
\qquad
\hat y=\arg\max_{y'\in\mathcal Y}p_\phi(y'\mid x,z_{1:K}).
\label{eq:vlr-inference-v13}
\end{equation}
The intended computation therefore has two functional stages:
\begin{equation}
x
\xrightarrow[\text{latent formation}]{G_\theta}
z_{1:K},
\qquad
(x,z_{1:K})
\xrightarrow[\text{answer generation}]{p_\phi}
\hat y.
\label{eq:intended-chain-v13}
\end{equation}
The first stage should develop the problem's intermediate reasoning within the latent trajectory; the second should draw on that trajectory to produce the answer. These roles define two requirements.

\paragraph{C1: Latent Reliance.}
In the common question-visible architecture of Equation~\ref{eq:vlr-inference-v13}, the answer model conditions on both $x$ and $z_{1:K}$. The contextual route $x\rightarrow\hat y$ may therefore coexist with the latent-mediated route $x\rightarrow z_{1:K}\rightarrow\hat y$. Latent reliance does not require the answer model to discard $x$ or to predict from $z_{1:K}$ alone. It requires the latent sequence to remain an active determinant of the answer distribution when the question is held fixed.

For a question $x$, C1 requires that disabling the latent input changes the answer distribution:
\begin{equation*}
\tag{C1}
C_1(x)
:=
\mathbf 1\!\left[
p_\phi(\cdot\mid x,G_\theta(x))
\neq
p_\phi(\cdot\mid x,\varnothing_z)
\right]
=1.
\end{equation*}
Thus, with $x$ unchanged, C1 makes the latent branch $z_{1:K}\rightarrow\hat y$ load-bearing within the full predictive computation. It asks whether the answer model uses the latent input; C2 concerns what it represents.

\paragraph{C2: Reasoning Capability.}
The latent-construction stage must itself support the reasoning that transforms the question into the answer. Let $\mathcal T(x,y)\subseteq\mathcal T$ be the set of valid derivations from $x$ to $y$, where each derivation contains the non-trivial intermediate inferences required by the problem. Let $\widehat c(z)$ denote the textual derivation recovered from a latent trajectory by its prescribed decoder, held fixed during evaluation and given only $z$. Define
\begin{equation}
V_{\mathrm{reason}}(x,z,y)
=
\mathbf 1\!\left[\widehat c(z)\in\mathcal T(x,y)\right].
\label{eq:reason-validity-v13}
\end{equation}
For a question--answer pair $(x,y)$, C2 requires
\begin{equation*}
\tag{C2}
C_2(x,y)
:=
V_{\mathrm{reason}}\!\left(x,G_\theta(x),y\right)=1.
\end{equation*}
The trajectory $G_\theta(x)$ is produced for the current problem, and its decoded text is judged against the corresponding set $\mathcal T(x,y)$. This condition concerns the represented derivation rather than the final answer alone, and applies equally to continuous visual states and discrete optical codes.

Together, C1 and C2 align the two stages of Equation~\ref{eq:intended-chain-v13}: the answer must rely on the latent trajectory, and the trajectory must carry a valid reasoning process. We use their conjunction as the criterion for visual latent reasoning:
{\small
\begin{equation}
\operatorname{VLR}(x,y)
:=
\underbrace{C_1(x)}_{\text{latent reliance}}
\ \land\ 
\underbrace{C_2(x,y)}_{\text{reasoning capability}}.
\label{eq:vlr-v13}
\end{equation}
}

\subsection{Interventional Analysis of Continuous and Discrete Latent Reasoners}
\label{sec:interventional-analysis-v13}

We analyze representative \emph{continuous} and \emph{discrete} optical-character latent reasoners under a shared interventional framework, detailed with supporting measurements in Appendix~\ref{app:prelim-analysis-v13}.

\begin{figure}[!htbp]
    \centering
    \noindent\raisebox{-1.51544pt}[107.86446pt][1.51544pt]{%
\includegraphics[width=\linewidth,height=109.37990pt]{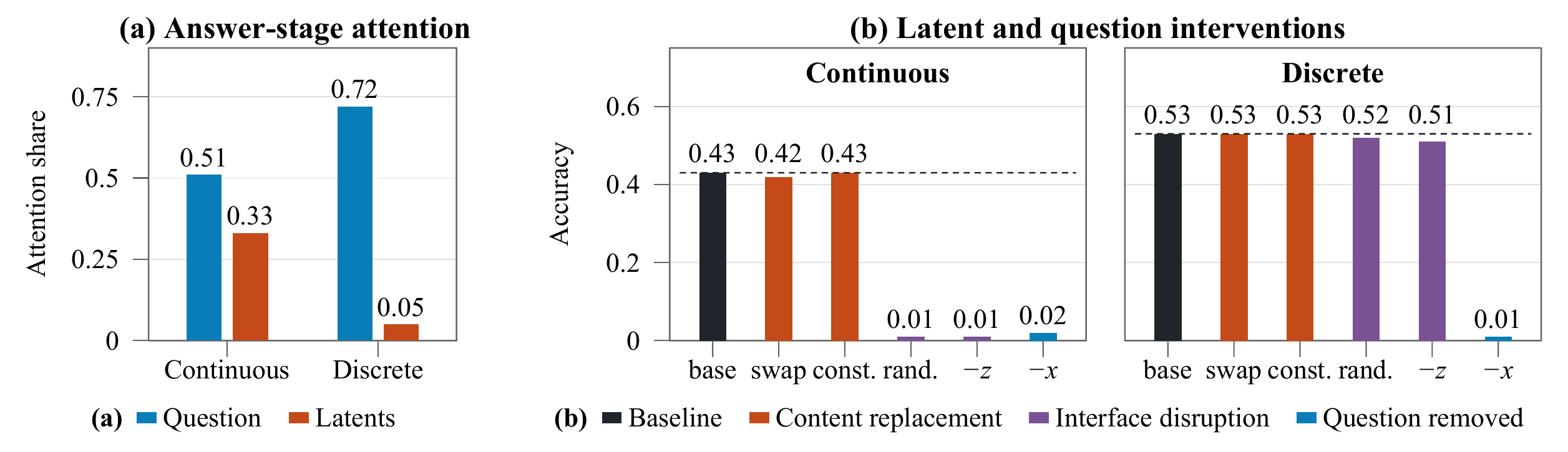}}\par
    \vspace{-10pt}
    \caption{\textbf{Complementary handoff failures in continuous and discrete latents.}
(a) Both favor the question, with higher latent attention in the continuous system.
(b) Content replacement has little effect ($\Lambda_{\mathrm{content}}\approx0$); interface disruption yields accuracy drops of $0.42$ and $0.015$, respectively.}
    \label{fig:prelim-interventions-v13}
\end{figure}

\paragraph{Latent interventions.}
With question and answer model fixed, we intervene on \emph{content} (cross-problem and constant replacement) or \emph{interface} (norm-matched randomization and deletion). For native accuracy $A_{\mathrm{base}}$ and intervened accuracy $A_\iota$, define $\Lambda_\iota=A_{\mathrm{base}}-A_\iota$; $\Lambda_{\mathrm{content}}$ and $\Lambda_{\mathrm{interface}}$ average each pair. Deletion tests C1, content replacement tests sensitivity to problem-specific latents, and derivation decoding assesses C2. Removing only the answer-side question, retaining native latents, gives $\Lambda_x=A_{\mathrm{base}}-A_{-x}$. Figure~\ref{fig:prelim-interventions-v13} shows results, Appendix~\ref{app:analysis-aggregation} defines the measures.

\paragraph{Continuous latents: a load-bearing interface without problem-specific reasoning.}
Removing or randomizing RoT's latent interface reduces accuracy from $0.43$ to $0.01$, whereas cross-example and constant replacement retain $0.42$ and $0.43$. Answer prediction thus relies on the interface (C1; $\Lambda_{\mathrm{interface}}=0.42$) but is insensitive to its problem-specific content; latent attention remains substantial despite higher question attention (Appendix~\ref{app:prelim-analysis-v13}). Yet decoding recovers no chain of thought from native states: they are interchangeable control signals, not problem-specific reasoning, failing C2.

\paragraph{Discrete latents: supervised codes bypassed by the answer model.}
DLR-SFT stays near its $0.53$ baseline under all four latent interventions: cross-example and constant replacement preserve accuracy; randomization and deletion yield $0.52$ and $0.51$. Removing the answer-side question reduces accuracy to $0.01$. Thus, the answer model bypasses supervised latents and relies on the question, failing C1, consistent with question-dominated attention and weak latent gradients.

These failures are complementary: the continuous system satisfies C1 but fails C2, the discrete system fails C1. Effective visual latent reasoning requires both.

\FloatBarrier

\FloatBarrier
\subsection{Question-Visible Training Decouples Latent Reasoning from Answering}
\label{sec:objective-account-v13}

The complementary failures point beyond representation format to the objective shared by both paradigms. Let $\mathcal D_{\mathrm{tr}}$ be a training distribution over triples $(x,c,y)$, where $c\in\mathcal T(x,y)$ is a reference derivation. A question-visible objective has the general form
\begin{equation}
\mathcal L_{\mathrm{vis}}(\theta,\phi)
=
\mathbb E_{(x,c,y)\sim\mathcal D_{\mathrm{tr}}}
\left[
-\log p_\phi\!\left(y\mid x,G_\theta(x)\right)
+
\lambda_{\mathrm{repr}}\mathcal L_{\mathrm{repr}}\!\left(G_\theta(x),c\right)
\right],
\label{eq:visible-objective-v13}
\end{equation}
where $\lambda_{\mathrm{repr}}\geq0$, and $\mathcal L_{\mathrm{repr}}$ may align continuous states with rendered features or make discrete codes reconstruct a reference derivation. The representation term shapes what $z$ represents; the answer term rewards the correct output while exposing both $x$ and $z$. Neither requires problem-specific reasoning to be both formed in $z$ and used for the answer.

This ambiguity is captured by the latent-content-invariant family
\begin{equation}
\mathcal P_{\mathrm{inv}}(\theta)
=
\left\{
p_\phi:
p_\phi(\cdot\mid x,z)=p_\phi(\cdot\mid x,z'),
\ \forall x\in\operatorname{supp}(\mathcal D_X),
\ \forall z,z'\in\mathcal Z_\theta
\right\},
\label{eq:invariant-family-v13}
\end{equation}
where $\mathcal Z_\theta=\{G_\theta(x):x\in\operatorname{supp}(\mathcal D_X)\}$ is the set of reachable trajectories. A question-only answer distribution belongs to this family. The representation term supervises $G_\theta$ without requiring the answer model to use its output, so representation supervision and latent-content-invariant answer prediction can coexist. Within this family, an answer model may still react to the presence of the latent interface, as in RoT, or bypass it entirely, as in DLR-SFT.

The objective therefore leaves latent formation and answer prediction uncoupled. Addressing C1 and C2 jointly calls for supervision that makes the same trajectory support answer prediction and encode a valid, problem-specific derivation.
\section{CALR: Functional Anchoring of Latent Reasoning}
\label{sec:method-v13}

\begin{figure}[!t]
\centering
\includegraphics[width=\linewidth]{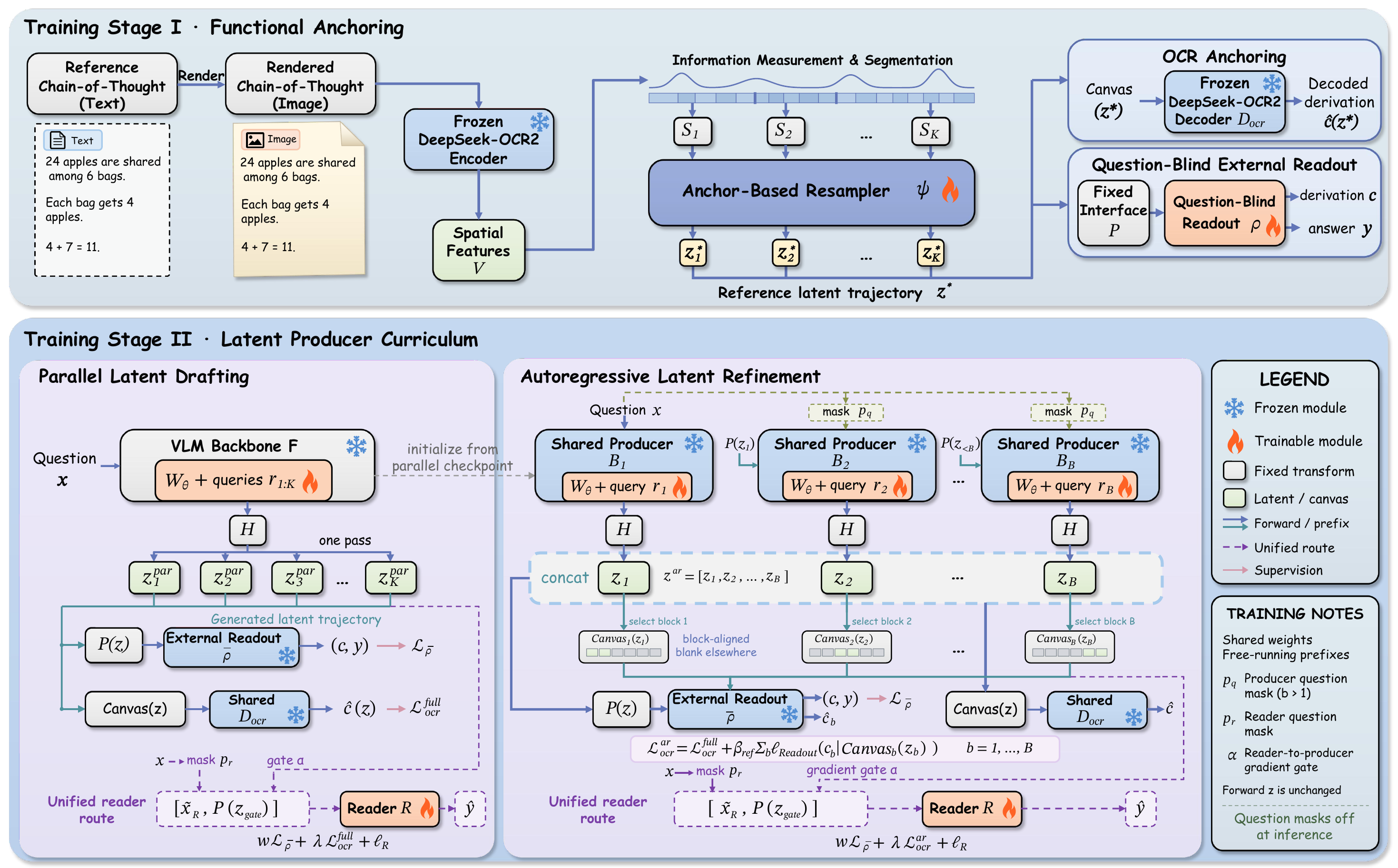}
\vspace{-15pt}
\caption{CALR training framework: functional anchoring followed by parallel drafting and autoregressive refinement. Purple dashed paths denote the unified internal-reader route.}
\label{fig:calr-framework}
\end{figure}

\subsection{Functional Formulation}
\label{sec:method-formulation-v15}

For $(x,c,y)\sim\mathcal D_{\mathrm{tr}}$ with $c\in\mathcal T(x,y)$, let $z=G_\theta(x)$. We formulate the supervision shared by both realizations through decoupled CALR. During producer learning, its question-blind reader $\bar\rho$, optical decoder $D_{\mathrm{ocr}}$, and input mappings $P$ and $\operatorname{Canvas}$ remain fixed. The answer model and prescribed decoder of Section~\ref{sec:criterion-v13} are instantiated as
\begin{equation}
p_\phi(y\mid x,z)
:=p_{\bar\rho}\!\left(y\mid P(z),\tau_{\mathrm{ans}}\right),
\qquad
\widehat c(z)
:=\operatorname{Decode}_{D_{\mathrm{ocr}}}\!\left(
\operatorname{Canvas}(z)\right).
\label{eq:functional-readouts-v15}
\end{equation}
The reader's mode tokens $\tau_{\mathrm{ans}}$ and $\tau_{\mathrm{cot}}$ select answer and derivation recovery, respectively. Both readouts receive the same trajectory without direct question input.

\paragraph{Joint supervision of answer use and reasoning content.}
Writing $\ell_M(t\mid u)=-\log p_M(t\mid u)$, the full-trajectory producer objective is
\begin{equation}
\min_\theta\ \mathbb E_{(x,c,y)\sim\mathcal D_{\mathrm{tr}}}\Big[
\underbrace{\ell_{\bar\rho}\!\left(y\mid P(z),\tau_{\mathrm{ans}}\right)}_{\text{answer supervision}}
+\underbrace{\eta\ell_{\bar\rho}\!\left(c\mid P(z),\tau_{\mathrm{cot}}\right)
+\lambda\ell_{D_{\mathrm{ocr}}}\!\left(c\mid\operatorname{Canvas}(z)\right)}_{\text{derivation supervision}}
\Big].
\label{eq:parallel-objective-v13}
\end{equation}
Here $\eta,\lambda\geq0$, $\mathcal L_{\bar\rho}(z)$ combines the reader's answer and $\eta$-weighted derivation losses, and $\mathcal L_{\mathrm{ocr}}^{\mathrm{full}}(z):=\ell_{D_{\mathrm{ocr}}}(c\mid\operatorname{Canvas}(z))$, giving $\mathcal L_{\bar\rho}(z)+\lambda\mathcal L_{\mathrm{ocr}}^{\mathrm{full}}(z)$ per example.

\paragraph{Coupling formation to use.}
Unlike Equation~\ref{eq:visible-objective-v13}, the answer term receives problem-specific information only through $G_\theta(x)$. A readout invariant across reachable trajectories would be question-independent, unable to exploit the direct question-conditioned route admitted by $\mathcal P_{\mathrm{inv}}(\theta)$. Latent-mediated answer supervision thus addresses the use requirement in C1. For C2, the OCR term supervises $c\in\mathcal T(x,y)$ through the prescribed evaluation decoder; the reader's derivation mode additionally supervises recovery through $P(z)$.

During producer learning, gradients pass through the fixed readouts and input mappings to update $G_\theta$ via $z$, adapting the trajectory under stable answer and derivation readout rules.

\subsection{Constructing Functional Anchors}
\label{sec:method-anchor-v13}

Stage~I establishes the reference trajectories and fixed readouts required by the objective before learning question-conditioned generation in Stage~II (Figure~\ref{fig:calr-framework}).

\paragraph{Information-balanced reference trajectories.}
To preserve derivation content within a fixed latent budget, we balance information across states. The frozen DeepSeek-OCR2 encoder $E$ maps a rendered derivation $c$ to spatial features $V\in\mathbb R^{N\times d}$. Corpus statistics $(\mu_j,\sigma_j)$, fixed before training, define an unnormalized diagonal-Gaussian feature-surprisal score $a_i$. We form $K$ contiguous segments of approximately equal cumulative score and resample each with global page context:
{\small
\begin{equation}
\begin{gathered}
V=E\!\left(\operatorname{render}(c)\right),\qquad
a_i=\frac{1}{2}\sum_{j=1}^{d}\left(\frac{V_{ij}-\mu_j}{\sigma_j}\right)^2,\qquad
\sum_{i\in S_g}a_i\approx\frac{1}{K}\sum_{i=1}^{N}a_i,\\
z_g^\star=\operatorname{Attn}_{\psi}\!\left(q_{\mathrm{pool}},\operatorname{Ctx}_{\psi}(V)_{S_g}\right),
\qquad g=1,\ldots,K.
\end{gathered}
\label{eq:anchored-construction-v13}
\end{equation}
}
Here $\psi$ comprises the shared learned query $q_{\mathrm{pool}}$, context, and resampling parameters, and $z^\star=(z_1^\star,\ldots,z_K^\star)$ with $z_g^\star\in\mathbb R^{d_z}$. Information-dense regions receive narrower spatial segments, while sparse regions receive wider ones. Reference text is separately segmented at paragraph and punctuation boundaries; spans are assigned in reading order, with $c_g$ denoting the target for latent $g$.

\paragraph{Anchoring derivation content.}
The frozen DeepSeek-OCR2 decoder $D_{\mathrm{ocr}}$ supervises both the complete reference trajectory and its individual segments. The full canvas places all $K$ latents on a fixed visual grid; $\operatorname{Canvas}_g(z_g)$ retains only position $g$. Unused positions contain blank-page features. We learn the resampler by minimizing
{\small
{\small
\begin{equation}
\begin{aligned}
&\mathcal L_{\mathrm{anchor}}
=\ell_{D_{\mathrm{ocr}}}\!\left(
c\mid\operatorname{Canvas}(z^\star)\right)\\
&\quad+\beta\sum_{g=1}^{K}
\ell_{D_{\mathrm{ocr}}}\!\left(
c_g\mid\operatorname{Canvas}_g(z_g^\star)\right)
+\gamma\Omega(z^\star),
&\Omega(z^\star)
=\frac{1}{K}\sum_{g=1}^{K}
\left(\lVert z_g^\star\rVert_2-\nu_0\right)^2,
\end{aligned}
\label{eq:anchor-objective-v13}
\end{equation}
}
}

where $\beta,\gamma\geq0$ and $\nu_0$ is the target feature norm. Global supervision preserves the complete derivation, while segment supervision makes its ordered parts individually decodable. These reference trajectories establish the representation on which the answer reader is trained. Appendix~\ref{app:anchoring-v13} details spatial segmentation, text targets, and canvas construction.

\paragraph{Anchoring answer recovery.}
The reader $\rho$ uses a trainable low-rank readout adapter and mode-token embeddings on a frozen base VLM $F$. It receives reference latents through a fixed scaled-isometric interface. For $d_z\leq d_{\mathrm{emb}}$, define
\begin{equation}
P=sU,\qquad H=s^{-1}U^\top,\qquad
U\in\mathbb R^{d_{\mathrm{emb}}\times d_z},\qquad
U^\top U=I_{d_z},\qquad HP=I_{d_z},
\label{eq:fixed-interface-v13}
\end{equation}
where $s>0$. Both maps act position-wise: $P$ embeds optical latents in the VLM input space, and $H$ maps producer states back to optical coordinates. Keeping the resampler and reference latents fixed, we train $\rho$ without the question:
\begin{equation}
\mathcal L_{\mathrm{contract}}
=\ell_{\rho}\!\left(y\mid P(z^\star),\tau_{\mathrm{ans}}\right)
+\eta\ell_{\rho}\!\left(c\mid P(z^\star),\tau_{\mathrm{cot}}\right).
\label{eq:contract-objective-v13}
\end{equation}
The answer and derivation targets make both functions accessible from the anchored representation. We then freeze $\rho$ as $\bar\rho$. Stage~II learns to generate trajectories that support these established readouts, with $P$ and $H$ also fixed throughout.

\subsection{Learning Anchored Latent Trajectories}
\label{sec:method-trajectories-v15}

Stage~II replaces reference trajectories from $c$ with trajectories generated from $x$. We train a parallel producer under the joint objective, then refine it autoregressively through prefix conditioning and block-aligned derivation recovery. Both phases support decoupled and unified CALR.

\paragraph{Parallel drafting.}
\label{sec:method-parallel-v13}
The producer uses the frozen base VLM $F$ with a low-rank route $W_\theta$ and learned queries $r_{1:K}$. These parameters and the special-token embeddings form $\theta$. Selecting query states with $\Pi_r$, we generate all $K$ latents in one pass:
\begin{equation}
z^{\mathrm{par}}=G_\theta^{\mathrm{par}}(x)
=H\!\left(\Pi_rF^{W_\theta}\!\left([x;r_{1:K}]\right)\right).
\label{eq:parallel-producer-v13}
\end{equation}
Training minimizes $\mathcal L_{\bar\rho}(z^{\mathrm{par}})+\lambda\mathcal L_{\mathrm{ocr}}^{\mathrm{full}}(z^{\mathrm{par}})$. This phase learns question-conditioned trajectories under the fixed readout functions before introducing inter-block feedback.

\paragraph{Autoregressive refinement.}
\label{sec:method-autoregressive-v13}
Warm-starting from the parallel-trained producer, we make later blocks condition on the generated prefix. Let $(\mathcal B_1,\ldots,\mathcal B_B)$ partition the $K$ latent positions into ordered contiguous blocks, with $\sum_b|\mathcal B_b|=K$. Write $\mathbf z_b:=z_{\mathcal B_b}$, $\mathbf r_b:=r_{\mathcal B_b}$, and $c_{\mathcal B_b}:=\operatorname{concat}_{g\in\mathcal B_b}c_g$ for latent blocks, query blocks, and their derivation targets. The recurrence is
\begin{equation}
\begin{gathered}
\mathbf z_b=H\!\left(\Pi_{\mathbf r_b}F^{W_\theta}\!\left(
[\widetilde x_b;P(\mathbf z_{<b});\mathbf r_b]\right)\right),\quad b=1,\ldots,B; \ \
z^{\mathrm{ar}}=\operatorname{concat}(\mathbf z_1,\ldots,\mathbf z_B).
\end{gathered}
\label{eq:autoregressive-producer-v13}
\end{equation}
Here $\mathbf z_{<b}$ concatenates preceding blocks and $\widetilde x_1=x$. For $b\geq2$, $\widetilde x_b$ is replaced by $\langle\mathrm{blind}\rangle$ with probability $p_q$; on these steps, problem-specific information must reach the next block through the prefix. Training and inference use free-running prefixes, with gradients connected across blocks during training.

Prefix conditioning governs inter-block information flow, while the frozen external reader supervises each block's reasoning content. We add block-aligned recovery to full-trajectory OCR supervision:
{\small
\begin{equation}
\mathcal L_{\mathrm{ocr}}^{\mathrm{ar}}(z^{\mathrm{ar}})
=\mathcal L_{\mathrm{ocr}}^{\mathrm{full}}(z^{\mathrm{ar}})
+\beta_{\mathrm{ref}}\sum_{b=1}^{B}
\ell_{\mathrm{Read}_{\mathrm{out}}}\!\left(
c_{\mathcal B_b}\mid
\operatorname{Canvas}_{\mathcal B_b}(\mathbf z_b)
\right).
\label{eq:autoregressive-objective-v13}
\end{equation}
}

Here $\mathrm{Read}_{\mathrm{out}}$ denotes the canvas-based derivation readout of $\bar\rho$, with its fixed input interface and $\tau_{\mathrm{cot}}$ implicit. Block canvases retain absolute positions with blank-page padding. We minimize $\mathcal L_{\bar\rho}(z^{\mathrm{ar}})+\lambda\mathcal L_{\mathrm{ocr}}^{\mathrm{ar}}(z^{\mathrm{ar}})$, retaining full-trajectory reader losses for answers and derivations.

\paragraph{Unified CALR.}
\label{sec:method-realizations-v13}
Unified CALR learns answer use inside the \emph{same VLM} $F$ while retaining the fixed external supervision. An internal reader adapter $R$, initialized from the Stage~I readout adapter, defines the answer model $p_R:=p_{F^R}$ and remains trainable alongside the separate producer adapter. It receives $P(z)$ and $\widetilde x_R$, which equals $\langle\mathrm{blind}\rangle$ with probability $p_r$ and $x$ otherwise. This reader mask trains answer recovery from latents even when question context is absent. We separate learning to read the trajectory from influencing its generation through a gradient gate:
\begin{equation}
\begin{aligned}
z_{\mathrm{gate}}^m
&=\alpha z^m+(1-\alpha)\operatorname{sg}[z^m],\qquad\alpha\in[0,1],\\
\mathcal L_{\mathrm{uni}}^m
&=w\mathcal L_{\bar\rho}(z^m)+\lambda\mathcal L_{\mathrm{ocr}}^m(z^m)
+\mathbb E_{\widetilde x_R}\ell_R\!\left(y\mid\widetilde x_R,P(z_{\mathrm{gate}}^m),\tau_{\mathrm{ans}}\right).
\end{aligned}
\label{eq:unified-objective-v13}
\end{equation}
Here $m\in\{\mathrm{par},\mathrm{ar}\}$, $\mathcal L_{\mathrm{ocr}}^{\mathrm{par}}:=\mathcal L_{\mathrm{ocr}}^{\mathrm{full}}$, $w\geq w_{\min}>0$, and $\operatorname{sg}$ stops gradients. The gate preserves $z_{\mathrm{gate}}^m=z^m$ in the forward pass and scales only the internal reader's gradient into the producer; $R$ continues learning at $\alpha=0$. The fixed external losses remain active throughout. At inference, both masks are disabled: decoupled CALR uses the frozen external reader on $P(z^{\mathrm{ar}})$, while unified CALR uses the internal reader on $[x;P(z^{\mathrm{ar}})]$. Appendix~\ref{app:calr-realizations} details gradient admission and inference.
\section{Experiments}
\label{sec:experiments-v14}

\subsection{Experimental Setup}
\label{sec:experimental-setup}

\paragraph{Training and base model.}
We train the reader and producer on GSM8K-Aug. Resampler training additionally uses rendered reference derivations from MATH and MathX. The frozen DeepSeek-OCR2 encoder and decoder supply optical features and supervision, separately from the base VLM. Our main model is Qwen3-VL-4B-Instruct, with evaluations also covering Qwen3-VL at 2B and 8B and LLaVA-V1.6-Mistral-7B. CALR uses 32 continuous latents, arranged as four blocks of eight during autoregressive refinement. Appendix~\ref{app:experimental-setup} details data allocation and training configurations.

\paragraph{Evaluation and metrics.}
We evaluate on GSM8K-Aug, GSM-Hard, SVAMP, MultiArith, and MATH-500, with disjoint training and evaluation data. Models receive only the question at inference. We report greedy-decoding final-answer accuracy (Pass@1 or Acc.) and intermediate reasoning length (\#L), using unweighted benchmark averages. Appendix~\ref{app:latent-analysis} details diagnostics and ablations.

\subsection{Main Results}
\label{sec:main-results}

\begingroup
\setlength{\intextsep}{8pt plus 2pt minus 2pt}
\setlength{\textfloatsep}{10pt plus 2pt minus 2pt}
\begin{table}[!htbp]
\centering
\caption{Results on four grade-school reasoning benchmarks with Qwen3-VL-4B-Instruct. Par.\ and AR denote parallel drafting and autoregressive refinement, respectively.}
\label{tab:main-results-v14}
\begingroup
\fontsize{8}{10}\selectfont
\setlength{\tabcolsep}{1.8pt}
\renewcommand{\arraystretch}{1.06}
\newcommand{\MRMethodLabel}{Method}
\newcommand{\MRAverageLabel}{Average}
\sbox{\MRtablebox}{%
\begin{tabular}{@{}l*{10}{c}@{}}
\toprule
& \multicolumn{2}{c}{\textbf{GSM8K-Aug}}
& \multicolumn{2}{c}{\textbf{GSM-Hard}}
& \multicolumn{2}{c}{\textbf{SVAMP}}
& \multicolumn{2}{c}{\textbf{MultiArith}}
& \multicolumn{2}{c}{\textbf{\MRAverageLabel}}\\
\cmidrule(lr){2-3}\cmidrule(lr){4-5}\cmidrule(lr){6-7}\cmidrule(lr){8-9}\cmidrule(l){10-11}
\textbf{\MRMethodLabel} & Acc. & \#L & Acc. & \#L & Acc. & \#L & Acc. & \#L & Acc. & \#L\\
\midrule
\rowcolor{MRgroup}
\multicolumn{11}{l}{\strut\bfseries\itshape Qwen3-VL-4B-Instruct (LMM)}\\
Direct-SFT & \MRcell{26.2}{.3}{} & \MRcell{0.0}{.0}{} & \MRcell{9.5}{.1}{} & \MRcell{0.0}{.0}{} & \MRcell{70.0}{.3}{} & \MRcell{0.0}{.0}{} & \MRcell{85.6}{.4}{} & \MRcell{0.0}{.0}{} & $47.8$ & $0.0$\\
CoT-SFT & \MRcell{81.2}{.4}{} & \MRcell{127.3}{1.0}{} & \MRcell{53.4}{.3}{} & \MRcell{191.1}{1.6}{} & \MRcell{84.3}{.3}{} & \MRcell{55.9}{.6}{} & \MRcell{98.3}{.4}{} & \MRcell{59.1}{.6}{} & $79.3$ & $108.4$\\
\midrule
Coconut (CoLM 2025) & \MRcell{16.1}{.4}{} & \MRcell{6.0}{.0}{} & \MRcell{5.2}{.3}{} & \MRcell{6.0}{.0}{} & \MRcell{42.0}{.3}{} & \MRcell{6.0}{.0}{} & \MRcell{58.9}{.2}{} & \MRcell{6.0}{.0}{} & $30.6$ & $6.0$\\
CODI (EMNLP 2025) & \MRcell{7.1}{.3}{} & \MRcell{6.0}{.0}{} & \MRcell{2.0}{.4}{} & \MRcell{6.0}{.0}{} & \MRcell{10.7}{.1}{} & \MRcell{6.0}{.0}{} & \MRcell{17.8}{.2}{} & \MRcell{6.0}{.0}{} & $9.4$ & $6.0$\\
CoLaR (NeurIPS 2025) & \MRcell{39.2}{.2}{} & \MRcell{38.8}{.5}{} & \MRcell{8.2}{.4}{} & \MRcell{47.8}{.7}{} & \MRcell{52.0}{.4}{} & \MRcell{18.4}{.6}{} & \MRcell{81.0}{.3}{} & \MRcell{22.1}{.3}{} & $45.1$ & $31.8$\\
RoT (ACL 2026) & \MRcell{37.8}{.3}{} & \MRcell{32.0}{.0}{} & \MRcell{14.1}{.2}{} & \MRcell{32.0}{.0}{} & \MRcell{72.7}{.4}{} & \MRcell{32.0}{.0}{} & \MRcell{97.2}{.4}{} & \MRcell{32.0}{.0}{} & $55.5$ & $32.0$\\
DLR (SFT) & \MRcell{48.0}{.4}{} & \MRcell{6.3}{.4}{} & \MRcell{14.7}{.3}{} & \MRcell{6.8}{.5}{} & \MRcell{68.7}{.2}{} & \MRcell{3.6}{.1}{} & \MRcell{97.8}{.3}{} & \MRcell{4.2}{.2}{} & $57.3$ & $5.2$\\
\midrule
\rowcolor{MRdecoupled}
\textit{Ours, decoupled (Par.)} & \MRcell{51.2}{.2}{} & \MRcell{32.0}{.0}{} & \MRcell{13.8}{.3}{} & \MRcell{32.0}{.0}{} & \MRcell{71.3}{.2}{} & \MRcell{32.0}{.0}{} & \MRcell{97.6}{.1}{} & \MRcell{32.0}{.0}{} & $58.5$ & $32.0$\\
\rowcolor{MRunified}
\textit{Ours, unified (Par.)} & \MRcell{52.4}{.3}{} & \MRcell{32.0}{.0}{} & \MRcell{14.3}{.2}{} & \MRcell{32.0}{.0}{} & \MRcell{72.4}{.1}{} & \MRcell{32.0}{.0}{} & \MRcell{98.1}{.1}{} & \MRcell{32.0}{.0}{} & $59.3$ & $32.0$\\
\rowcolor{MRdecoupled}
\textit{Ours, decoupled (AR)} & \MRcell{62.1}{.2}{} & \MRcell{32.0}{.0}{} & \MRcell{16.1}{.2}{} & \MRcell{32.0}{.0}{} & \MRcell{76.6}{.3}{} & \MRcell{32.0}{.0}{} & \MRcell{98.8}{.1}{} & \MRcell{32.0}{.0}{} & $63.4$ & $32.0$\\
\rowcolor{MRunified}
\textit{Ours, unified (AR)} & \MRcell{\MRbest{63.8}}{.2}{} & \MRcell{32.0}{.0}{} & \MRcell{\MRbest{16.8}}{.3}{} & \MRcell{32.0}{.0}{} & \MRcell{\MRbest{77.4}}{.1}{} & \MRcell{32.0}{.0}{} & \MRcell{\MRbest{99.7}}{.1}{} & \MRcell{32.0}{.0}{} & $\MRbest{64.4}$ & $32.0$\\
\bottomrule
\end{tabular}
\relax{}}
\ifdim\wd\MRtablebox>\linewidth
\resizebox{\linewidth}{!}{\usebox{\MRtablebox}}%
\else
\usebox{\MRtablebox}%
\fi
\endgroup
\vspace{-15pt}
\end{table}
\endgroup
\paragraph{Supervised latent reasoning.}
Table~\ref{tab:main-results-v14} compares CALR with supervised latent-reasoning baselines on Qwen3-VL-4B-Instruct. Unified autoregressive CALR achieves the highest accuracy among the compared latent methods on all four benchmarks, averaging 64.4\% versus 57.3\% for DLR (SFT). Autoregressive refinement benefits both realizations: average accuracy rises from 58.5\% to 63.4\% for decoupled CALR and from 59.3\% to 64.4\% for unified CALR, with all configurations retaining 32 latents. Unified autoregressive CALR further improves on its decoupled counterpart on every benchmark. These results show that continuous latent reasoning can outperform discrete latent coding under supervised training, with refinement improving both external and internal readout realizations. The complete comparison appears in Appendix~\ref{app:extended-results}.

\begingroup
\setlength{\intextsep}{8pt plus 2pt minus 2pt}
\setlength{\textfloatsep}{10pt plus 2pt minus 2pt}
\begin{table}[!htb]
\centering
\caption{Comparison with RoT across three vision-language models on four mathematical reasoning benchmarks. Pass@1 is reported in percent, and \#L denotes reasoning length.}
\label{tab:rot-comparison}
\begingroup
\fontsize{8}{10}\selectfont
\setlength{\tabcolsep}{1.6pt}
\renewcommand{\arraystretch}{1.06}
\newcommand{\MRMethodLabel}{Method}
\newcommand{\MRAverageLabel}{Average}
\sbox{\MRtablebox}{%
\begin{tabular}{@{}l@{\hspace{5pt}}*{11}{c}@{}}
\toprule
& \multicolumn{2}{c}{\textbf{GSM8K-Aug}}
& \multicolumn{2}{c}{\textbf{GSM-Hard}}
& \multicolumn{2}{c}{\textbf{SVAMP}}
& \multicolumn{2}{c}{\textbf{MultiArith}}
& \multicolumn{3}{c}{\textbf{\MRAverageLabel}}\\
\cmidrule(lr){2-3}\cmidrule(lr){4-5}\cmidrule(lr){6-7}\cmidrule(lr){8-9}\cmidrule(l){10-12}
\textbf{\MRMethodLabel} & Pass@1 & \#L & Pass@1 & \#L & Pass@1 & \#L & Pass@1 & \#L & Pass@1 & \#L & Pass@1/\#L\\
\midrule
\rowcolor{MRgroup}
\multicolumn{12}{l}{\strut\bfseries\itshape Qwen3-VL-2B-Instruct}\\
SFT-w/o & \MRcell{15.6}{.3}{} & \MRcell{0.0}{.0}{} & \MRcell{4.7}{.2}{} & \MRcell{0.0}{.0}{} & \MRcell{52.3}{.3}{} & \MRcell{0.0}{.0}{} & \MRcell{41.7}{.3}{} & \MRcell{0.0}{.0}{} & $28.6$ & $0.0$ & --\\
SFT-CoT & \MRcell{59.7}{.4}{} & \MRcell{131.4}{1.6}{} & \MRcell{33.1}{.3}{} & \MRcell{207.2}{1.7}{} & \MRcell{67.3}{.3}{} & \MRcell{63.4}{.8}{} & \MRcell{95.0}{.4}{} & \MRcell{68.0}{.7}{} & $63.8$ & $117.5$ & $0.54$\\
RoT & \MRcell{23.3}{.3}{} & \MRcell{32.0}{.0}{} & \MRcell{8.6}{.2}{} & \MRcell{32.0}{.0}{} & \MRcell{53.7}{.4}{} & \MRcell{32.0}{.0}{} & \MRcell{62.2}{.4}{} & \MRcell{32.0}{.0}{} & $37.0$ & $32.0$ & $1.16$\\
\rowcolor{MRunified}
\textbf{Ours (CALR)} & \MRcell{42.8}{.3}{} & \MRcell{32.0}{.0}{} & \MRcell{10.6}{.2}{} & \MRcell{32.0}{.0}{} & \MRcell{59.5}{.3}{} & \MRcell{32.0}{.0}{} & \MRcell{85.5}{.2}{} & \MRcell{32.0}{.0}{} & $49.6$ & $32.0$ & $\MRbest{1.55}$\\
\midrule
\rowcolor{MRgroup}
\multicolumn{12}{l}{\strut\bfseries\itshape Qwen3-VL-4B-Instruct}\\
SFT-w/o & \MRcell{26.2}{.3}{} & \MRcell{0.0}{.0}{} & \MRcell{9.5}{.1}{} & \MRcell{0.0}{.0}{} & \MRcell{70.0}{.3}{} & \MRcell{0.0}{.0}{} & \MRcell{85.6}{.4}{} & \MRcell{0.0}{.0}{} & $47.8$ & $0.0$ & --\\
SFT-CoT & \MRcell{81.2}{.4}{} & \MRcell{127.3}{1.0}{} & \MRcell{53.4}{.3}{} & \MRcell{191.1}{1.6}{} & \MRcell{84.3}{.3}{} & \MRcell{55.9}{.6}{} & \MRcell{98.3}{.4}{} & \MRcell{59.1}{.6}{} & $79.3$ & $108.4$ & $0.73$\\
RoT & \MRcell{37.8}{.3}{} & \MRcell{32.0}{.0}{} & \MRcell{14.1}{.2}{} & \MRcell{32.0}{.0}{} & \MRcell{72.7}{.4}{} & \MRcell{32.0}{.0}{} & \MRcell{97.2}{.4}{} & \MRcell{32.0}{.0}{} & $55.5$ & $32.0$ & $1.73$\\
\rowcolor{MRunified}
\textbf{Ours (CALR)} & \MRcell{63.8}{.2}{} & \MRcell{32.0}{.0}{} & \MRcell{16.8}{.3}{} & \MRcell{32.0}{.0}{} & \MRcell{77.4}{.1}{} & \MRcell{32.0}{.0}{} & \MRcell{99.7}{.1}{} & \MRcell{32.0}{.0}{} & $64.4$ & $32.0$ & $\MRbest{2.01}$\\
\midrule
\rowcolor{MRgroup}
\multicolumn{12}{l}{\strut\bfseries\itshape LLaVA-V1.6-Mistral-7B}\\
SFT-w/o & \MRcell{10.8}{.2}{} & \MRcell{0.0}{.0}{} & \MRcell{2.3}{.1}{} & \MRcell{0.0}{.0}{} & \MRcell{40.7}{.3}{} & \MRcell{0.0}{.0}{} & \MRcell{52.8}{.5}{} & \MRcell{0.0}{.0}{} & $26.7$ & $0.0$ & --\\
SFT-CoT & \MRcell{38.6}{.3}{} & \MRcell{151.3}{1.6}{} & \MRcell{12.2}{.1}{} & \MRcell{209.7}{1.8}{} & \MRcell{56.3}{.5}{} & \MRcell{79.9}{.9}{} & \MRcell{86.1}{.5}{} & \MRcell{82.3}{.8}{} & $48.3$ & $130.8$ & $0.37$\\
RoT & \MRcell{16.3}{.2}{} & \MRcell{32.0}{.0}{} & \MRcell{3.6}{.1}{} & \MRcell{32.0}{.0}{} & \MRcell{49.0}{.4}{} & \MRcell{32.0}{.0}{} & \MRcell{68.3}{.5}{} & \MRcell{32.0}{.0}{} & $34.3$ & $32.0$ & $1.07$\\
\rowcolor{MRunified}
\textbf{Ours (CALR)} & \MRcell{28.4}{.3}{} & \MRcell{32.0}{.0}{} & \MRcell{5.2}{.2}{} & \MRcell{32.0}{.0}{} & \MRcell{52.7}{.2}{} & \MRcell{32.0}{.0}{} & \MRcell{81.3}{.3}{} & \MRcell{32.0}{.0}{} & $41.9$ & $32.0$ & $\MRbest{1.31}$\\
\bottomrule
\end{tabular}
\relax{}}
\ifdim\wd\MRtablebox>\linewidth
\resizebox{\linewidth}{!}{\usebox{\MRtablebox}}%
\else
\usebox{\MRtablebox}%
\fi
\endgroup
\vspace{-10pt}
\end{table}
\endgroup
\paragraph{Consistency across vision-language models.}
Table~\ref{tab:rot-comparison} compares CALR with RoT at the same length of 32 continuous latents. CALR improves Pass@1 in all twelve model--benchmark combinations: average accuracy rises from 37.0\% to 49.6\% on Qwen3-VL-2B, from 55.5\% to 64.4\% on Qwen3-VL-4B, and from 34.3\% to 41.9\% on LLaVA-Mistral-7B. The GSM8K-Aug gain on Qwen3-VL-4B reaches 26.0 percentage points. CALR also attains the highest reported Pass@1/\#L in each model group. These gains span model families and scales without increasing latent sequence length.

\begin{table}[!htb]
\centering
\begingroup
\setlength{\abovecaptionskip}{3pt}
\newcommand{\MCCaption}{Qwen3-VL on MATH-500.}
\newcommand{\MCLabel}{tab:math500-results}
\newcommand{\CABCaption}{Unified training ablations.}
\newcommand{\CABLabel}{tab:component-ablation}
\newcommand{\MRMethodLabel}{Method}
\newcommand{\CABBody}{tables/component_ablation}
\newsavebox{\CABLeftCaptionBox}
\newsavebox{\CABRightCaptionBox}
\newlength{\CABCaptionHeight}
\sbox{\CABLeftCaptionBox}{\begin{minipage}[t]{.555\linewidth}
  \expandafter\caption\expandafter{\MCCaption}\label{\MCLabel}
\end{minipage}}
\sbox{\CABRightCaptionBox}{\begin{minipage}[t]{.435\linewidth}
  \expandafter\caption\expandafter{\CABCaption}\label{\CABLabel}
\end{minipage}}
\setlength{\CABCaptionHeight}{\dimexpr\ht\CABLeftCaptionBox+\dp\CABLeftCaptionBox\relax}
\ifdim\dimexpr\ht\CABRightCaptionBox+\dp\CABRightCaptionBox\relax>\CABCaptionHeight
  \setlength{\CABCaptionHeight}{\dimexpr\ht\CABRightCaptionBox+\dp\CABRightCaptionBox\relax}
\fi
\begin{minipage}[t]{.555\linewidth}\vspace{0pt}
  \parbox[t][\CABCaptionHeight][t]{\linewidth}{\usebox{\CABLeftCaptionBox}}\par
  \centering
  \fontsize{8}{10}\selectfont
  \setlength{\tabcolsep}{1pt}
  \renewcommand{\arraystretch}{1.06}
  \newcommand{\MRScale}[1]{#1}
  \newcommand{\MROursLabel}{CALR}
  \sbox{\MRtablebox}{%
\providecommand{\MRScale}[1]{Qwen3-VL-#1}
\providecommand{\MROursLabel}{Ours (CALR)}
\begin{tabular}{@{}l*{6}{c}@{}}
\toprule
& \multicolumn{2}{c}{\textbf{\MRScale{2B}}}
& \multicolumn{2}{c}{\textbf{\MRScale{4B}}}
& \multicolumn{2}{c}{\textbf{\MRScale{8B}}}\\
\cmidrule(lr){2-3}\cmidrule(lr){4-5}\cmidrule(l){6-7}
\textbf{\MRMethodLabel} & Acc. & \#L & Acc. & \#L & Acc. & \#L\\
\midrule
Direct-SFT & \MRcell{20.8}{.2}{} & \MRcell{0.0}{.0}{} & \MRcell{29.4}{.3}{} & \MRcell{0.0}{.0}{} & \MRcell{33.5}{.5}{} & \MRcell{0.0}{.0}{}\\
CoT-SFT & \MRcell{29.2}{.3}{} & \MRcell{324.5}{2.6}{} & \MRcell{55.8}{.4}{} & \MRcell{291.5}{1.9}{} & \MRcell{69.4}{.3}{} & \MRcell{310.6}{1.7}{}\\
\midrule
RoT & \MRcell{24.0}{.2}{} & \MRcell{64.0}{.0}{} & \MRcell{33.2}{.4}{} & \MRcell{64.0}{.0}{} & \MRcell{37.8}{.3}{} & \MRcell{64.0}{.0}{}\\
DLR (SFT) & \MRcell{29.4}{.3}{} & \MRcell{26.7}{.8}{} & \MRcell{35.4}{.3}{} & \MRcell{27.9}{.7}{} & \MRcell{39.8}{.4}{} & \MRcell{27.2}{.5}{}\\
DLR (RL) & \MRcell{34.0}{.3}{} & \MRcell{28.9}{.6}{} & \MRcell{\MRbest{44.6}}{.4}{} & \MRcell{28.4}{.5}{} & \MRcell{54.0}{.3}{} & \MRcell{29.5}{.6}{}\\
\midrule
\rowcolor{MRunified}
\textbf{\MROursLabel} & \MRcell{\MRbest{34.6}}{.2}{} & \MRcell{32.0}{.0}{} & \MRcell{43.3}{.3}{} & \MRcell{32.0}{.0}{} & \MRcell{\MRbest{55.7}}{.2}{} & \MRcell{32.0}{.0}{}\\
\bottomrule
\end{tabular}
\relax{}}
  \ifdim\wd\MRtablebox>\linewidth
    \resizebox{\linewidth}{!}{\usebox{\MRtablebox}}
  \else
    \usebox{\MRtablebox}
  \fi
\end{minipage}\hfill
\begin{minipage}[t]{.435\linewidth}\vspace{0pt}
  \parbox[t][\CABCaptionHeight][t]{\linewidth}{\usebox{\CABRightCaptionBox}}\par
  \centering
  \fontsize{8}{10}\selectfont
  \setlength{\tabcolsep}{1.25pt}
  \renewcommand{\arraystretch}{1.06}
  \sbox{\MRtablebox}{%
\newcommand{\CABCurriculum}{Training}
\newcommand{\CABComponents}{Training components}
\newcommand{\CABExternal}{Ext.}
\newcommand{\CABReaderMask}{Mask}
\newcommand{\CABParallel}{Par. only}
\newcommand{\CABAutoregressive}{Par. $\rightarrow$ AR}
\begin{tabular}{@{}lccccc@{}}
\toprule
\multirow{2}{*}{\CABCurriculum}
 & \multicolumn{3}{c}{\CABComponents}
 & \multicolumn{2}{c}{\textbf{GSM8K-Aug}}\\
\cmidrule(lr){2-4}\cmidrule(l){5-6}
 & \CABExternal~(Par.)
 & \CABReaderMask~($p_r$)
 & \CABExternal~(AR)
 & Pass@1 & \#L\\
\midrule
\multirow{3}{*}{\CABParallel}
 & $\times$ & $\times$ & \textemdash & 39.7 & 32.0\\
 & $\checkmark$ & $\times$ & \textemdash & 46.8 & 32.0\\
 & $\checkmark$ & $\checkmark$ & \textemdash & 52.2 & 32.0\\
\midrule
\multirow{3}{*}{\CABAutoregressive}
 & $\checkmark$ & $\times$ & $\times$ & 53.7 & 32.0\\
 & $\checkmark$ & $\checkmark$ & $\times$ & 56.3 & 32.0\\
 & \cellcolor{MRunified}$\checkmark$
 & \cellcolor{MRunified}$\checkmark$
 & \cellcolor{MRunified}$\checkmark$
 & \cellcolor{MRunified}$\MRbest{63.8}$
 & \cellcolor{MRunified}32.0\\
\bottomrule
\end{tabular}
\relax{}}
  \ifdim\wd\MRtablebox>\linewidth
    \resizebox{\linewidth}{!}{\usebox{\MRtablebox}}
  \else
    \usebox{\MRtablebox}
  \fi
\end{minipage}
\endgroup
\vspace{-5pt}
\end{table}
\paragraph{Mathematical reasoning across scales.}
On MATH-500 (Table~\ref{tab:math500-results}), CALR outperforms RoT at all three Qwen3-VL scales while using 32 rather than 64 latents. Its advantage over DLR (SFT) increases from 5.2 points at 2B to 7.9 at 4B and 15.9 at 8B, showing that the continuous formulation benefits from larger models in this setting. Without reinforcement learning, CALR also attains the highest latent-method accuracy at 2B and 8B; at 4B, it reaches 43.3\%, compared with 44.6\% for DLR (RL). Together, these results extend CALR's accuracy gains beyond the grade-school benchmarks.

\Needspace{12\baselineskip}
\subsection{Component Ablations}
\label{sec:component-ablations}
\begingroup
\setlength{\intextsep}{6pt plus 1pt minus 1pt}
\setlength{\textfloatsep}{8pt plus 1pt minus 1pt}
\setlength{\columnsep}{10pt}
\textbf{Training curriculum and supervision.}
External supervision contributes at both stages of the curriculum (Table~\ref{tab:component-ablation}). In parallel training, it improves Pass@1 from 39.7\% to 46.8\%, and reader question masking further raises it to 52.2\%. During autoregressive refinement with masking, removing external supervision yields 56.3\%, compared with 63.8\% when retained. This 7.5-point gap demonstrates the value of continued external supervision after the parallel warm start. With both components, refinement improves over parallel training by 11.6 points while retaining 32 latents. These results support autoregressive refinement with latent-mediated supervision and reader question masking.
\par

\Needspace{12\baselineskip}
\begin{wraptable}[8]{r}{.40\linewidth}
\vspace{-8pt}
\centering
\setlength{\abovecaptionskip}{3pt}
\caption{Resampler ablations.}
\label{tab:resampler-ablation}
\begingroup
\fontsize{9}{11}\selectfont
\setlength{\tabcolsep}{2pt}
\renewcommand{\arraystretch}{1.06}
\newcommand{\RABResampling}{Resampling}
\newcommand{\RABSimilarity}{Sim. (\%)}
\newcommand{\RABGlobal}{Global cross-attn.}
\newcommand{\RABFixed}{Fixed anchors}
\newcommand{\RABDynamic}{Dynamic anchors}
\begin{tabular}{@{}lcc@{}}
\toprule
\multirow{2}{*}{\RABResampling} & \multicolumn{2}{c}{\textbf{GSM8K-Aug}}\\
\cmidrule(l){2-3}
 & \RABSimilarity\,$\uparrow$ & CE\,$\downarrow$\\
\midrule
\RABGlobal & 92.8 & 0.183\\
\RABFixed & 95.3 & 0.158\\
\rowcolor{MRunified}
\RABDynamic & $\MRbest{99.1}$ & $\MRbest{0.107}$\\
\bottomrule
\end{tabular}
\endgroup
\end{wraptable}
\textbf{Information-balanced resampling.}
Table~\ref{tab:resampler-ablation} links reconstruction fidelity to the resampling strategy. Fixed anchors improve on global cross-attention, but ignore variation in information density. Adapting segment widths to information content raises character similarity from 95.3\% to 99.1\% and reduces decoder cross-entropy from 0.158 to 0.107. Balancing information per latent thus preserves reference derivations more faithfully than assigning equal feature counts. Settings and metrics are detailed in Appendix~\ref{app:component-ablations}.
Complementary diagnostics (Appendix~\ref{sec:latent-analysis}) show that cross-problem latent swaps reduce CALR's accuracy from 64.3\% to 1.4\% with the question retained, while decoded trajectories recover 74\% of intermediate values for the current problem versus 7\% for others. Feedback interventions further support inter-block dependence.
\FloatBarrier
\endgroup

\section{Conclusion}
\label{sec:conclusion}

In this paper, we propose CALR to bridge the gap between latent formation and answer prediction in render-of-thought compression. Guided by latent reliance and reasoning capability, CALR couples latent-mediated answer supervision with derivation anchoring. A parallel-to-autoregressive curriculum develops continuous trajectories whose successive blocks build on generated prefixes. Experiments across five mathematical reasoning benchmarks and multiple model families demonstrate improved accuracy with compact latent states. Interventions and derivation decoding show that these states both support answer prediction and carry problem-specific intermediate reasoning. Together, these findings establish functional anchoring as a means of jointly shaping what continuous latents represent and how they support reasoning.
\section{Related Work}
\label{sec:related-work}

\paragraph{Latent reasoning.}
Continuous methods internalize reasoning through hidden-state recurrence in Coconut~\citep{hao2025coconut}, self-distillation in CODI~\citep{shen2025codi}, and dynamic trace compression in CoLaR~\citep{tan2025colar}. SoftCoT~\citep{xu2025softcot} supplies assistant-generated soft thoughts. SIM-CoT~\citep{wei2026simcot} adds step-aligned supervision through a training-only auxiliary decoder. CALR couples derivation anchoring with question-blind answer supervision, jointly shaping what latents represent and how they support prediction. See Appendix~\ref{app:related-work} for more.

\bibliography{references}

\begin{thebibliography}{41}
\providecommand{\natexlab}[1]{#1}
\providecommand{\url}[1]{\texttt{#1}}
\expandafter\ifx\csname urlstyle\endcsname\relax
  \providecommand{\doi}[1]{doi: #1}\else
  \providecommand{\doi}{doi: \begingroup \urlstyle{rm}\Url}\fi

\bibitem[Bai et~al.(2025)Bai, Cai, Chen, Chen, Chen, Cheng,
  et~al.]{bai2025qwen3vl}
Shuai Bai, Yuxuan Cai, Ruizhe Chen, Keqin Chen, Xionghui Chen, Zesen Cheng,
  et~al.
\newblock {Qwen3-VL} technical report.
\newblock \emph{arXiv preprint arXiv:2511.21631}, 2025.
\newblock URL \url{https://arxiv.org/abs/2511.21631}.

\bibitem[Chang et~al.(2026)Chang, Liu, Wang, Gao, Ma, Zhao, Miao, Sun, Peng,
  Li, and Niu]{chang2026dlr}
Shuochen Chang, Qingyang Liu, Shaobo Wang, Bingjie Gao, Qianli Ma, Haonan Zhao,
  Yibo Miao, Yulin Sun, Zelin Peng, Jiangtong Li, and Li~Niu.
\newblock Why struggle with continuous latents? interpretable discrete latent
  reasoning via rendered compression.
\newblock \emph{arXiv preprint arXiv:2606.29712}, 2026.
\newblock URL \url{https://arxiv.org/abs/2606.29712}.

\bibitem[Chen et~al.(2025)Chen, Benton, Radhakrishnan, Uesato, Denison,
  Schulman, Somani, Hase, Wagner, Roger, Mikulik, Bowman, Leike, Kaplan, and
  Perez]{chen2025reasoning}
Yanda Chen, Joe Benton, Ansh Radhakrishnan, Jonathan Uesato, Carson Denison,
  John Schulman, Arushi Somani, Peter Hase, Misha Wagner, Fabien Roger, Vlad
  Mikulik, Samuel~R. Bowman, Jan Leike, Jared Kaplan, and Ethan Perez.
\newblock Reasoning models don't always say what they think.
\newblock \emph{arXiv preprint arXiv:2505.05410}, 2025.
\newblock URL \url{https://arxiv.org/abs/2505.05410}.

\bibitem[Cheng et~al.(2026)Cheng, Liu, Zhang, Fei, Hong, Lyu, Wang, Su, Gu,
  Liu, Bai, Tang, Wang, and Huang]{cheng2026glyph}
Jiale Cheng, Yusen Liu, Xinyu Zhang, Yulin Fei, Wenyi Hong, Ruiliang Lyu,
  Weihan Wang, Zhe Su, Xiaotao Gu, Xiao Liu, Yushi Bai, Jie Tang, Hongning
  Wang, and Minlie Huang.
\newblock {Glyph}: Scaling context windows via visual-text compression.
\newblock In \emph{Proceedings of the 64th Annual Meeting of the Association
  for Computational Linguistics (Volume 1: Long Papers)}, pp.\  37145--37158.
  Association for Computational Linguistics, 2026.
\newblock \doi{10.18653/v1/2026.acl-long.1722}.
\newblock URL \url{https://aclanthology.org/2026.acl-long.1722/}.

\bibitem[Cobbe et~al.(2021)Cobbe, Kosaraju, Bavarian, Chen, Jun, Kaiser,
  Plappert, Tworek, Hilton, Nakano, Hesse, and Schulman]{cobbe2021gsm8k}
Karl Cobbe, Vineet Kosaraju, Mohammad Bavarian, Mark Chen, Heewoo Jun, Lukasz
  Kaiser, Matthias Plappert, Jerry Tworek, Jacob Hilton, Reiichiro Nakano,
  Christopher Hesse, and John Schulman.
\newblock Training verifiers to solve math word problems.
\newblock \emph{arXiv preprint arXiv:2110.14168}, 2021.
\newblock URL \url{https://arxiv.org/abs/2110.14168}.

\bibitem[Deng et~al.(2023)Deng, Prasad, Fernandez, Smolensky, Chaudhary, and
  Shieber]{deng2023implicit}
Yuntian Deng, Kiran Prasad, Roland Fernandez, Paul Smolensky, Vishrav
  Chaudhary, and Stuart Shieber.
\newblock Implicit chain of thought reasoning via knowledge distillation.
\newblock \emph{arXiv preprint arXiv:2311.01460}, 2023.
\newblock URL \url{https://arxiv.org/abs/2311.01460}.

\bibitem[Deng et~al.(2024)Deng, Choi, and Shieber]{deng2024internalize}
Yuntian Deng, Yejin Choi, and Stuart Shieber.
\newblock From explicit {CoT} to implicit {CoT}: Learning to internalize {CoT}
  step by step.
\newblock \emph{arXiv preprint arXiv:2405.14838}, 2024.
\newblock URL \url{https://arxiv.org/abs/2405.14838}.

\bibitem[Dilgren \& Wiegreffe(2026)Dilgren and
  Wiegreffe]{dilgren2026interpretable}
Connor Dilgren and Sarah Wiegreffe.
\newblock Are latent reasoning models easily interpretable?
\newblock In \emph{Conference on Language Modeling}, 2026.
\newblock URL \url{https://arxiv.org/abs/2604.04902}.

\bibitem[Elazar et~al.(2021)Elazar, Ravfogel, Jacovi, and
  Goldberg]{elazar2021amnesic}
Yanai Elazar, Shauli Ravfogel, Alon Jacovi, and Yoav Goldberg.
\newblock Amnesic probing: Behavioral explanation with amnesic counterfactuals.
\newblock \emph{Transactions of the Association for Computational Linguistics},
  9:\penalty0 160--175, 2021.
\newblock \doi{10.1162/tacl_a_00359}.
\newblock URL \url{https://aclanthology.org/2021.tacl-1.10/}.

\bibitem[Gao et~al.(2023)Gao, Madaan, Zhou, Alon, Liu, Yang, Callan, and
  Neubig]{gao2023pal}
Luyu Gao, Aman Madaan, Shuyan Zhou, Uri Alon, Pengfei Liu, Yiming Yang, Jamie
  Callan, and Graham Neubig.
\newblock {PAL}: Program-aided language models.
\newblock In \emph{Proceedings of the 40th International Conference on Machine
  Learning}, volume 202 of \emph{Proceedings of Machine Learning Research},
  pp.\  10764--10799. PMLR, 2023.
\newblock URL \url{https://proceedings.mlr.press/v202/gao23f.html}.

\bibitem[Gozeten et~al.(2026)Gozeten, Ildiz, Zhang, Harutyunyan, Rawat, and
  Oymak]{gozeten2026cot2}
Halil~Alperen Gozeten, M.~Emrullah Ildiz, Xuechen Zhang, Hrayr Harutyunyan,
  Ankit~Singh Rawat, and Samet Oymak.
\newblock Continuous chain of thought enables parallel exploration and
  reasoning.
\newblock In \emph{International Conference on Learning Representations}, 2026.
\newblock URL \url{https://arxiv.org/abs/2505.23648}.

\bibitem[Hao et~al.(2025)Hao, Sukhbaatar, Su, Li, Hu, Weston, and
  Tian]{hao2025coconut}
Shibo Hao, Sainbayar Sukhbaatar, DiJia Su, Xian Li, Zhiting Hu, Jason Weston,
  and Yuandong Tian.
\newblock Training large language models to reason in a continuous latent
  space.
\newblock In \emph{Conference on Language Modeling}, 2025.
\newblock URL \url{https://openreview.net/forum?id=Itxz7S4Ip3}.

\bibitem[Hendrycks et~al.(2021)Hendrycks, Burns, Kadavath, Arora, Basart, Tang,
  Song, and Steinhardt]{hendrycks2021math}
Dan Hendrycks, Collin Burns, Saurav Kadavath, Akul Arora, Steven Basart, Eric
  Tang, Dawn Song, and Jacob Steinhardt.
\newblock Measuring mathematical problem solving with the {MATH} dataset.
\newblock In \emph{Proceedings of the Neural Information Processing Systems
  Track on Datasets and Benchmarks}, volume~1, 2021.
\newblock URL \url{https://arxiv.org/abs/2103.03874}.

\bibitem[Hu et~al.(2022)Hu, Shen, Wallis, Allen-Zhu, Li, Wang, Wang, and
  Chen]{hu2022lora}
Edward~J. Hu, Yelong Shen, Phillip Wallis, Zeyuan Allen-Zhu, Yuanzhi Li, Shean
  Wang, Lu~Wang, and Weizhu Chen.
\newblock {LoRA}: Low-rank adaptation of large language models.
\newblock In \emph{International Conference on Learning Representations}, 2022.
\newblock URL \url{https://openreview.net/forum?id=nZeVKeeFYf9}.

\bibitem[Jain \& Wallace(2019)Jain and Wallace]{jain2019attention}
Sarthak Jain and Byron~C. Wallace.
\newblock Attention is not explanation.
\newblock In \emph{Proceedings of the 2019 Conference of the North American
  Chapter of the Association for Computational Linguistics: Human Language
  Technologies, Volume 1 (Long and Short Papers)}, pp.\  3543--3556.
  Association for Computational Linguistics, 2019.
\newblock \doi{10.18653/v1/N19-1357}.
\newblock URL \url{https://aclanthology.org/N19-1357/}.

\bibitem[Jiang et~al.(2026)Jiang, Fu, Fang, Gao, Wang, He, and
  Li]{jiang2026univlr}
Houcheng Jiang, Jiajun Fu, Junfeng Fang, Chen Gao, Xiang Wang, Xiangnan He, and
  Yong Li.
\newblock {UniVLR}: Unifying text and vision in visual latent reasoning for
  multimodal {LLMs}.
\newblock \emph{arXiv preprint arXiv:2605.11856}, 2026.
\newblock URL \url{https://arxiv.org/abs/2605.11856}.

\bibitem[Lanham et~al.(2023)Lanham, Chen, Radhakrishnan, Steiner, Denison,
  Hernandez, Li, Durmus, Hubinger, Kernion, Luko{\v s}i{\={u}}t{\.{e}}, Nguyen,
  Cheng, Joseph, Schiefer, Rausch, Larson, McCandlish, Kundu, Kadavath, Yang,
  Henighan, Maxwell, Telleen-Lawton, Hume, Hatfield-Dodds, Kaplan, Brauner,
  Bowman, and Perez]{lanham2023faithfulness}
Tamera Lanham, Anna Chen, Ansh Radhakrishnan, Benoit Steiner, Carson Denison,
  Danny Hernandez, Dustin Li, Esin Durmus, Evan Hubinger, Jackson Kernion,
  Kamil{\.{e}} Luko{\v s}i{\={u}}t{\.{e}}, Karina Nguyen, Newton Cheng,
  Nicholas Joseph, Nicholas Schiefer, Oliver Rausch, Robin Larson, Sam
  McCandlish, Sandipan Kundu, Saurav Kadavath, Shannon Yang, Thomas Henighan,
  Timothy Maxwell, Timothy Telleen-Lawton, Tristan Hume, Zac Hatfield-Dodds,
  Jared Kaplan, Jan Brauner, Samuel~R. Bowman, and Ethan Perez.
\newblock Measuring faithfulness in chain-of-thought reasoning.
\newblock \emph{arXiv preprint arXiv:2307.13702}, 2023.
\newblock URL \url{https://arxiv.org/abs/2307.13702}.

\bibitem[Li et~al.(2026)Li, Chen, Ju, Yao, Li, Lan, Zhu, Zheng, and
  Wang]{li2026pmps}
Yiqi Li, Xu~Chen, Chen Ju, Jiangchao Yao, Zhaoyang Li, Jinsong Lan, Xiaoyong
  Zhu, Bo~Zheng, and Yu~Wang.
\newblock Structural process supervision for latent chain-of-thought reasoning.
\newblock \emph{arXiv preprint arXiv:2609.09928}, 2026.
\newblock URL \url{https://arxiv.org/abs/2609.09928}.

\bibitem[Lightman et~al.(2024)Lightman, Kosaraju, Burda, Edwards, Baker, Lee,
  Leike, Schulman, Sutskever, and Cobbe]{lightman2024verify}
Hunter Lightman, Vineet Kosaraju, Yura Burda, Harri Edwards, Bowen Baker, Teddy
  Lee, Jan Leike, John Schulman, Ilya Sutskever, and Karl Cobbe.
\newblock Let's verify step by step.
\newblock In \emph{The Twelfth International Conference on Learning
  Representations}, 2024.
\newblock URL \url{https://openreview.net/forum?id=v8L0pN6EOi}.

\bibitem[Liu et~al.(2024)Liu, Li, Li, Li, Zhang, Shen, and
  Lee]{liu2024llavanext}
Haotian Liu, Chunyuan Li, Yuheng Li, Bo~Li, Yuanhan Zhang, Sheng Shen, and
  Yong~Jae Lee.
\newblock {LLaVA-NeXT}: Improved reasoning, {OCR}, and world knowledge, January
  2024.
\newblock URL \url{https://llava-vl.github.io/blog/2024-01-30-llava-next/}.

\bibitem[Loshchilov \& Hutter(2019)Loshchilov and Hutter]{loshchilov2019adamw}
Ilya Loshchilov and Frank Hutter.
\newblock Decoupled weight decay regularization.
\newblock In \emph{International Conference on Learning Representations}, 2019.
\newblock URL \url{https://openreview.net/forum?id=Bkg6RiCqY7}.

\bibitem[Mangrulkar et~al.(2022)Mangrulkar, Gugger, Debut, Belkada, Paul,
  Bossan, and Tietz]{mangrulkar2022peft}
Sourab Mangrulkar, Sylvain Gugger, Lysandre Debut, Younes Belkada, Sayak Paul,
  Benjamin Bossan, and Marian Tietz.
\newblock {PEFT}: State-of-the-art parameter-efficient fine-tuning methods,
  2022.
\newblock URL \url{https://github.com/huggingface/peft}.

\bibitem[Paszke et~al.(2019)Paszke, Gross, Massa, Lerer, Bradbury, Chanan,
  Killeen, Lin, Gimelshein, Antiga, Desmaison, Kopf, Yang, DeVito, Raison,
  Tejani, Chilamkurthy, Steiner, Fang, Bai, and Chintala]{paszke2019pytorch}
Adam Paszke, Sam Gross, Francisco Massa, Adam Lerer, James Bradbury, Gregory
  Chanan, Trevor Killeen, Zeming Lin, Natalia Gimelshein, Luca Antiga, Alban
  Desmaison, Andreas Kopf, Edward Yang, Zachary DeVito, Martin Raison, Alykhan
  Tejani, Sasank Chilamkurthy, Benoit Steiner, Lu~Fang, Junjie Bai, and Soumith
  Chintala.
\newblock {PyTorch}: An imperative style, high-performance deep learning
  library.
\newblock In \emph{Advances in Neural Information Processing Systems},
  volume~32, 2019.
\newblock URL
  \url{https://papers.nips.cc/paper_files/paper/2019/hash/bdbca288fee7f92f2bfa9f7012727740-Abstract.html}.

\bibitem[Patel et~al.(2021)Patel, Bhattamishra, and Goyal]{patel2021svamp}
Arkil Patel, Satwik Bhattamishra, and Navin Goyal.
\newblock Are {NLP} models really able to solve simple math word problems?
\newblock In \emph{Proceedings of the 2021 Conference of the North American
  Chapter of the Association for Computational Linguistics: Human Language
  Technologies}, pp.\  2080--2094. Association for Computational Linguistics,
  2021.
\newblock \doi{10.18653/v1/2021.naacl-main.168}.
\newblock URL \url{https://aclanthology.org/2021.naacl-main.168/}.

\bibitem[Rawal(2024)]{rawal2024mathx}
Parvesh Rawal.
\newblock {MathX}: Large-scale mathematical reasoning dataset.
\newblock Hugging Face dataset, 2024.
\newblock URL \url{https://huggingface.co/datasets/Modotte/MathX-5M}.
\newblock Released by Modotte.

\bibitem[Roy \& Roth(2015)Roy and Roth]{roy2015multiarith}
Subhro Roy and Dan Roth.
\newblock Solving general arithmetic word problems.
\newblock In \emph{Proceedings of the 2015 Conference on Empirical Methods in
  Natural Language Processing}, pp.\  1743--1752. Association for Computational
  Linguistics, 2015.
\newblock \doi{10.18653/v1/D15-1202}.
\newblock URL \url{https://aclanthology.org/D15-1202/}.

\bibitem[Rust et~al.(2023)Rust, Lotz, Bugliarello, Salesky, de~Lhoneux, and
  Elliott]{rust2023pixel}
Phillip Rust, Jonas~F. Lotz, Emanuele Bugliarello, Elizabeth Salesky, Miryam
  de~Lhoneux, and Desmond Elliott.
\newblock Language modelling with pixels.
\newblock In \emph{International Conference on Learning Representations}, 2023.
\newblock URL \url{https://arxiv.org/abs/2207.06991}.

\bibitem[Shen et~al.(2025)Shen, Yan, Zhang, Hu, Du, and He]{shen2025codi}
Zhenyi Shen, Hanqi Yan, Linhai Zhang, Zhanghao Hu, Yali Du, and Yulan He.
\newblock {CODI}: Compressing chain-of-thought into continuous space via
  self-distillation.
\newblock In \emph{Proceedings of the 2025 Conference on Empirical Methods in
  Natural Language Processing}, pp.\  677--693. Association for Computational
  Linguistics, 2025.
\newblock \doi{10.18653/v1/2025.emnlp-main.36}.
\newblock URL \url{https://aclanthology.org/2025.emnlp-main.36/}.

\bibitem[Tan et~al.(2025)Tan, Li, Ju, Luo, Song, and Luan]{tan2025colar}
Wenhui Tan, Jiaze Li, Jianzhong Ju, Zhenbo Luo, Ruihua Song, and Jian Luan.
\newblock Think silently, think fast: Dynamic latent compression of {LLM}
  reasoning chains.
\newblock In \emph{Advances in Neural Information Processing Systems},
  volume~38, 2025.
\newblock \doi{10.52202/085713-0164}.

\bibitem[Turpin et~al.(2023)Turpin, Michael, Perez, and
  Bowman]{turpin2023unfaithful}
Miles Turpin, Julian Michael, Ethan Perez, and Samuel~R. Bowman.
\newblock Language models don't always say what they think: Unfaithful
  explanations in chain-of-thought prompting.
\newblock In \emph{Advances in Neural Information Processing Systems},
  volume~36, pp.\  74952--74965, 2023.
\newblock URL \url{https://arxiv.org/abs/2305.04388}.

\bibitem[Wang et~al.(2026{\natexlab{a}})Wang, Liu, Zhao, Xu, and
  Gao]{wang2026regular}
Fanmeng Wang, Haotian Liu, Guojiang Zhao, Hongteng Xu, and Zhifeng Gao.
\newblock {ReGuLaR}: Variational latent reasoning guided by rendered
  chain-of-thought.
\newblock \emph{arXiv preprint arXiv:2601.23184}, 2026{\natexlab{a}}.
\newblock URL \url{https://arxiv.org/abs/2601.23184}.

\bibitem[Wang et~al.(2026{\natexlab{b}})Wang, Li, Li, Yang, Wei, and
  Tang]{wang2026rot}
Yifan Wang, Shiyu Li, Peiming Li, Xiaochen Yang, Zheng Wei, and Yang Tang.
\newblock {Render-of-Thought}: Rendering textual chain-of-thought as images for
  visual latent reasoning.
\newblock In \emph{Proceedings of the 64th Annual Meeting of the Association
  for Computational Linguistics (Volume 1: Long Papers)}, pp.\  45236--45253.
  Association for Computational Linguistics, 2026{\natexlab{b}}.
\newblock \doi{10.18653/v1/2026.acl-long.2097}.
\newblock URL \url{https://aclanthology.org/2026.acl-long.2097/}.

\bibitem[Wei et~al.(2025)Wei, Sun, and Li]{wei2025deepseekocr}
Haoran Wei, Yaofeng Sun, and Yukun Li.
\newblock {DeepSeek-OCR}: Contexts optical compression.
\newblock \emph{arXiv preprint arXiv:2510.18234}, 2025.
\newblock URL \url{https://arxiv.org/abs/2510.18234}.

\bibitem[Wei et~al.(2026{\natexlab{a}})Wei, Sun, and Li]{wei2026deepseekocr2}
Haoran Wei, Yaofeng Sun, and Yukun Li.
\newblock {DeepSeek-OCR 2}: Visual causal flow.
\newblock \emph{arXiv preprint arXiv:2601.20552}, 2026{\natexlab{a}}.
\newblock URL \url{https://arxiv.org/abs/2601.20552}.

\bibitem[Wei et~al.(2022)Wei, Wang, Schuurmans, Bosma, Ichter, Xia, Chi, Le,
  and Zhou]{wei2022cot}
Jason Wei, Xuezhi Wang, Dale Schuurmans, Maarten Bosma, Brian Ichter, Fei Xia,
  Ed~Chi, Quoc~V. Le, and Denny Zhou.
\newblock Chain-of-thought prompting elicits reasoning in large language
  models.
\newblock In \emph{Advances in Neural Information Processing Systems},
  volume~35, 2022.
\newblock \doi{10.52202/068431-1800}.
\newblock URL \url{https://arxiv.org/abs/2201.11903}.

\bibitem[Wei et~al.(2026{\natexlab{b}})Wei, Liu, Zang, Dong, Cao, Wang, Qiu,
  and Lin]{wei2026simcot}
Xilin Wei, Xiaoran Liu, Yuhang Zang, Xiaoyi Dong, Yuhang Cao, Jiaqi Wang,
  Xipeng Qiu, and Dahua Lin.
\newblock {SIM-CoT}: Supervised implicit chain-of-thought.
\newblock In \emph{International Conference on Learning Representations},
  2026{\natexlab{b}}.
\newblock URL \url{https://arxiv.org/abs/2509.20317}.

\bibitem[Wolf et~al.(2020)Wolf, Debut, Sanh, Chaumond, Delangue, Moi, Cistac,
  Rault, Louf, Funtowicz, Davison, Shleifer, von Platen, Ma, Jernite, Plu, Xu,
  Le~Scao, Gugger, Drame, Lhoest, and Rush]{wolf2020transformers}
Thomas Wolf, Lysandre Debut, Victor Sanh, Julien Chaumond, Clement Delangue,
  Anthony Moi, Pierric Cistac, Tim Rault, Remi Louf, Morgan Funtowicz, Joe
  Davison, Sam Shleifer, Patrick von Platen, Clara Ma, Yacine Jernite, Julien
  Plu, Canwen Xu, Teven Le~Scao, Sylvain Gugger, Mariama Drame, Quentin Lhoest,
  and Alexander Rush.
\newblock Transformers: State-of-the-art natural language processing.
\newblock In \emph{Proceedings of the 2020 Conference on Empirical Methods in
  Natural Language Processing: System Demonstrations}, pp.\  38--45.
  Association for Computational Linguistics, 2020.
\newblock \doi{10.18653/v1/2020.emnlp-demos.6}.
\newblock URL \url{https://aclanthology.org/2020.emnlp-demos.6/}.

\bibitem[Wu et~al.(2025)Wu, Teng, and Tu]{wu2025pccot}
Haoyi Wu, Zhihao Teng, and Kewei Tu.
\newblock Parallel continuous chain-of-thought with jacobi iteration.
\newblock In \emph{Proceedings of the 2025 Conference on Empirical Methods in
  Natural Language Processing}, pp.\  914--926. Association for Computational
  Linguistics, 2025.
\newblock \doi{10.18653/v1/2025.emnlp-main.47}.
\newblock URL \url{https://aclanthology.org/2025.emnlp-main.47/}.

\bibitem[Xu et~al.(2025{\natexlab{a}})Xu, Guo, Zeng, and Miao]{xu2025softcot}
Yige Xu, Xu~Guo, Zhiwei Zeng, and Chunyan Miao.
\newblock {SoftCoT}: Soft chain-of-thought for efficient reasoning with {LLMs}.
\newblock In \emph{Proceedings of the 63rd Annual Meeting of the Association
  for Computational Linguistics (Volume 1: Long Papers)}, pp.\  23336--23351.
  Association for Computational Linguistics, 2025{\natexlab{a}}.
\newblock \doi{10.18653/v1/2025.acl-long.1137}.
\newblock URL \url{https://aclanthology.org/2025.acl-long.1137/}.

\bibitem[Xu et~al.(2025{\natexlab{b}})Xu, Guo, Zeng, and Miao]{xu2025softcotpp}
Yige Xu, Xu~Guo, Zhiwei Zeng, and Chunyan Miao.
\newblock {SoftCoT++}: Test-time scaling with soft chain-of-thought reasoning.
\newblock \emph{arXiv preprint arXiv:2505.11484}, 2025{\natexlab{b}}.
\newblock URL \url{https://arxiv.org/abs/2505.11484}.

\bibitem[Yang et~al.(2025)Yang, Li, Yang, Zhang, Hui, Zheng,
  et~al.]{yang2025qwen3}
An~Yang, Anfeng Li, Baosong Yang, Beichen Zhang, Binyuan Hui, Bo~Zheng, et~al.
\newblock {Qwen3} technical report.
\newblock \emph{arXiv preprint arXiv:2505.09388}, 2025.
\newblock URL \url{https://arxiv.org/abs/2505.09388}.

\end{thebibliography}
\bibliographystyle{iclr2027_conference}

\clearpage

\appendix

\section{Extended Related Work}
\label{app:related-work}

\paragraph{Latent reasoning and process supervision.}
Early approaches internalize explicit derivations through hidden-state distillation and progressive removal of verbalized steps~\citep{deng2023implicit,deng2024internalize}. Subsequent work develops complementary ways to construct and learn latent trajectories. Coconut~\citep{hao2025coconut} feeds hidden states back as continuous thoughts, CODI~\citep{shen2025codi} transfers explicit reasoning through self-distillation, and CoLaR~\citep{tan2025colar} dynamically compresses reasoning traces. SoftCoT~\citep{xu2025softcot} supplies assistant-generated soft thoughts, while SoftCoT++~\citep{xu2025softcotpp} encourages diverse latent paths for test-time scaling. Parallelism takes different forms: PCCoT~\citep{wu2025pccot} updates latent positions through Jacobi iterations, whereas CoT2~\citep{gozeten2026cot2} represents multiple reasoning paths through mixtures of vocabulary embeddings. Beyond learning compact states, process supervision specifies the reasoning they should preserve. SIM-CoT~\citep{wei2026simcot} applies step-aligned supervision through a training-only auxiliary decoder; concurrent prototype-mediated process supervision (PMPS)~\citep{li2026pmps} aligns latent and explicit trajectories through shared prototypes and many-to-many assignments. CALR connects this representational objective to answer prediction: derivation anchoring supervises latent reasoning content, while a question-blind external reader directs answer supervision through the generated trajectory.

\paragraph{Optical compression and rendered reasoning.}
Rendering provides a visual interface for language, exemplified by PIXEL~\citep{rust2023pixel}, which learns from image patches of text. DeepSeek-OCR~\citep{wei2025deepseekocr}, DeepSeek-OCR2~\citep{wei2026deepseekocr2}, and Glyph~\citep{cheng2026glyph} develop optical representations for compactly encoding textual contexts. Visual latent reasoning extends this interface from context encoding to supervision of problem-conditioned reasoning. RoT~\citep{wang2026rot} aligns generated latents with rendered-CoT features, while DLR~\citep{chang2026dlr} constructs a discrete optical vocabulary for autoregressive modeling. UniVLR~\citep{jiang2026univlr} combines textual derivations and auxiliary visual evidence within a shared rendered workspace. These approaches connect visual encoding with compact reasoning representations. CALR establishes an information-balanced continuous interface by anchoring reference trajectories and their segments with a frozen optical decoder, then couples derivation recovery with latent-mediated answer supervision through a frozen external reader.

\paragraph{Faithfulness of latent reasoning.}
Faithfulness concerns the relationship between represented reasoning and the computations governing a prediction. Studies of explicit CoT examine this relationship through input-bias probes~\citep{turpin2023unfaithful}, interventions on reasoning traces~\citep{lanham2023faithfulness}, and tests of whether reasoning models disclose answer-influencing hints~\citep{chen2025reasoning}. Representation analysis makes a related distinction between information availability and use. Attention interventions~\citep{jain2019attention} assess prediction sensitivity to attention distributions, while Amnesic Probing~\citep{elazar2021amnesic} removes decodable properties to test their contribution to task behavior. Concurrent work on latent reasoning examines token necessity and recovers textual traces through vocabulary projections~\citep{dilgren2026interpretable}. CALR studies latent reliance and reasoning capability jointly across continuous and discrete visual reasoners. Answer-input interventions test the contribution of generated latents, optical decoding assesses their problem-specific reasoning content, and feedback-prefix interventions examine dependencies between successive latent blocks. This combination connects what latent trajectories encode with their roles in answer prediction and subsequent reasoning.
\clearpage
\section{CALR Realizations and Experimental Setup}
\label{app:experimental-setup}

\subsection{Decoupled and Unified CALR: Training and Inference}
\label{app:calr-realizations}
\label{app:anchor-training-v13}
\label{app:readout-admission-v13}
\label{app:prefix-v13}

This section brings together CALR's two realizations, their training objectives, the prefix interventions used to assess autoregressive dependence, and the resulting gradient-admission and inference procedures.

\paragraph{Model roles and initialization.}
For a question--derivation--answer triple $(x,c,y)$, the producer $G_\theta$ generates $K$ optical latents $z=(z_1,\ldots,z_K)$. The fixed interface $P$ embeds these latents into the base VLM's input space, while $H$ maps selected producer states back to optical coordinates. Stage~I first learns the resampler $\psi$ with the optical encoder $E$ and decoder $D_{\mathrm{ocr}}$ frozen. Holding the resulting reference latents $z^\star$ fixed, we train the question-blind reader $\rho$ through its low-rank adapter and mode-token embeddings. Its answer and derivation modes receive the same $P(z^\star)$, with respective mode tokens $\tau_{\mathrm{ans}}$ and $\tau_{\mathrm{cot}}$. The trained reader is then frozen as $\bar\rho$.

Decoupled CALR pairs $G_\theta$ with this external readout: answer prediction conditions on $P(z)$ alone. Unified CALR performs latent generation and answer generation within the \emph{same VLM} $F$, using separate low-rank producer and internal-reader adapters while keeping the pretrained parameters frozen. The internal reader $R$ is initialized from the Stage~I readout adapter and remains trainable during Stage~II; $p_R:=p_{F^R}$ denotes its conditional distribution. It receives both the question and $P(z)$, while $\bar\rho$ remains a frozen, question-blind supervisor. Both realizations learn a parallel producer and warm-start autoregressive refinement from it.

\paragraph{Training objectives and gradient flow.}
Let $m\in\{\mathrm{par},\mathrm{ar}\}$ index the producer phase. The external loss $\mathcal L_{\bar\rho}(z^m)$ supervises both answers and derivations from $P(z^m)$. The loss $\mathcal L_{\bar\rho}(z^m)$ combines full-trajectory answer and derivation supervision from the frozen external reader. In the parallel phase, $\mathcal L_{\mathrm{ocr}}^{\mathrm{par}}:=\mathcal L_{\mathrm{ocr}}^{\mathrm{full}}$ uses full-trajectory OCR supervision. Autoregressive refinement additionally supervises aligned derivation spans through the external reader's block-canvas readout, as defined in Equation~\ref{eq:autoregressive-objective-v13}.
\begin{equation*}
\begin{aligned}
z_{\mathrm{gate}}^m
&=\alpha z^m+(1-\alpha)\operatorname{sg}[z^m],\qquad\alpha\in[0,1],\\
\mathcal L_{\mathrm{uni}}^m
&=w\mathcal L_{\bar\rho}(z^m)+\lambda\mathcal L_{\mathrm{ocr}}^m(z^m)
+\mathbb E_{\widetilde x_R}\ell_R\!\left(y\mid\widetilde x_R,P(z_{\mathrm{gate}}^m),\tau_{\mathrm{ans}}\right).
\end{aligned}
\end{equation*}
Here $\ell_R(t\mid u)=-\log p_R(t\mid u)$ is the negative log-likelihood of target $t$ given conditioning input $u$, and $\operatorname{sg}$ is stop-gradient. The internal reader's question input $\widetilde x_R$ equals $\langle\mathrm{blind}\rangle$ with probability $p_r$ and $x$ otherwise. This readout mask is distinct from the producer mask $p_q$, which replaces the question for autoregressive blocks $b\geq2$; the first block always receives $x$.

The gate satisfies $z_{\mathrm{gate}}^m=z^m$ in the forward pass and scales only the internal-reader gradient into the producer by $\alpha$. The internal reader continues learning when $\alpha=0$. The parameters of $\bar\rho$ and $D_{\mathrm{ocr}}$ stay frozen, while differentiation through their latent inputs continues to update the producer. Autoregressive prefixes remain connected across blocks. These properties separate learning to read the trajectory from admitting the internal reader's additional influence on its generation.

\paragraph{Paired prefix interventions.}
We assess dependence on generated prefixes by intervening on the producer's feedback and regenerating the subsequent blocks. Let $B$ denote the number of ordered latent blocks, $\mathcal D_{\mathrm{eval}}$ a paired question--answer evaluation set, and $\mathcal Q_{\mathrm{eval}}$ its producer-question-mask distribution. A realization $\omega$ specifies $(\widetilde x_1,\ldots,\widetilde x_B)$, with $\widetilde x_1=x$; native and intervened rollouts share this realization. Write $z^{\mathrm{ar}}(x;\omega)=\operatorname{concat}(\mathbf z_1,\ldots,\mathbf z_B)$ for the native trajectory, where $\mathbf z_b$ contains the latents of block $b$. At a boundary $b\geq2$, the replacement prefixes are
\begin{equation}
\mathbf u_{<b}^{\mathrm{zero}}=\mathbf 0_{<b},\qquad
\mathbf u_{<b}^{\mathrm{swap}}=\mathbf z'_{<b},
\label{eq:prefix-replacements-v13}
\end{equation}
Here $z'$ is an unmodified, shape-matched trajectory from another problem. Because $P$ is fixed and linear, zeroing feedback embeddings is equivalent to applying $P$ to a zero latent prefix. Let $W_\theta$ denote the producer adapter, $\mathbf r_j$ the learned queries for block $j$, and $\Pi_{\mathbf r_j}$ their state selector. For $\mathcal S\in\{\mathrm{zero},\mathrm{swap}\}$, the remaining blocks are regenerated as
\begin{equation}
\widehat{\mathbf z}_{j\mid b}^{\mathcal S}
=H\!\left(\Pi_{\mathbf r_j}F^{W_\theta}\!\left(
[\widetilde x_j;P([\mathbf u_{<b}^{\mathcal S};\widehat{\mathbf z}_{b:j-1\mid b}^{\mathcal S}]);\mathbf r_j]
\right)\right),\qquad j=b,\ldots,B.
\label{eq:prefix-intervention-rollout-v13}
\end{equation}
The generated portion is empty at $j=b$ and accumulates newly generated blocks thereafter. The donor prefix remains fixed throughout regeneration. The final readout receives
\begin{equation}
\bar z_b^{\mathcal S}
:=[\mathbf z_{<b};\widehat{\mathbf z}_{b:B\mid b}^{\mathcal S}],
\qquad
\widehat{\mathbf z}_{b:B\mid b}^{\mathcal S}
:=\operatorname{concat}(\widehat{\mathbf z}_{b\mid b}^{\mathcal S},\ldots,\widehat{\mathbf z}_{B\mid b}^{\mathcal S}).
\label{eq:prefix-final-readout-v13}
\end{equation}
Thus, replacement acts on the feedback copy of the prefix: the readout retains its native prefix and receives the regenerated suffix. Question tokens and latent queries are unchanged within each pair. The accuracy difference therefore measures how preceding blocks shape subsequent generation.

\paragraph{Measuring prefix dependence.}
Training-time readiness checks use the frozen external readout, whose answer prediction is
\begin{equation}
\widehat y_{\bar\rho}(z)
:=\arg\max_{y'}p_{\bar\rho}(y'\mid P(z),\tau_{\mathrm{ans}}).
\label{eq:check-prediction-v13}
\end{equation}
The paired native and intervened accuracies are
\begin{equation}
\begin{aligned}
A_{\bar\rho}^{\mathrm{native}}
&=\mathbb E_{(x,y)\sim\mathcal D_{\mathrm{eval}},\,\omega\sim\mathcal Q_{\mathrm{eval}}}
\mathbf 1[\widehat y_{\bar\rho}(z^{\mathrm{ar}}(x;\omega))=y],\\
A_{\bar\rho,b}^{\mathcal S}
&=\mathbb E_{(x,y)\sim\mathcal D_{\mathrm{eval}},\,\omega\sim\mathcal Q_{\mathrm{eval}}}
\mathbf 1[\widehat y_{\bar\rho}(\bar z_b^{\mathcal S}(x;\omega))=y].
\end{aligned}
\label{eq:check-accuracies-v13}
\end{equation}
For $B>1$, their difference is averaged over all intervention boundaries:
{\small
\begin{equation}
\Lambda_{\mathcal S}^{\mathrm{ar}}
=\frac{1}{B-1}\sum_{b=2}^{B}
\left(A_{\bar\rho}^{\mathrm{native}}-A_{\bar\rho,b}^{\mathcal S}\right),
\qquad\mathcal S\in\{\mathrm{zero},\mathrm{swap}\},
\label{eq:autoregressive-effect-v13}
\end{equation}
}
Both empirical averages use the same examples, mask realizations, and donor assignments. Zero interventions measure reliance on the feedback path; donor-prefix interventions measure reliance on its problem-specific content. For training checks, $\mathcal D_{\mathrm{eval}}=\mathcal D_{\mathrm{check}}$ and $\mathcal Q_{\mathrm{eval}}=\mathcal Q_{\mathrm{check}}$ denote the fixed training-check set and its question-mask distribution.

\paragraph{Readiness-guided gradient admission.}
The parallel phase gradually introduces the internal reader's gradient into the producer. Autoregressive refinement resets $\alpha$ to zero, allowing the frozen supervision paths to first develop the blockwise producer. Let $\Lambda_{\mathcal S}^{\mathrm{ar}}(t)$ denote the effect at check $t$. Readiness requires both interventions to reach a positive threshold in two consecutive checks:
\begin{equation}
\min_{\mathcal S\in\{\mathrm{zero},\mathrm{swap}\}}
\min_{k\in\{t-1,t\}}\Lambda_{\mathcal S}^{\mathrm{ar}}(k)
\geq\delta_{\mathrm{dep}},\qquad\delta_{\mathrm{dep}}>0.
\label{eq:readout-admission-v13}
\end{equation}
We evaluate this criterion every 500 steps on a fixed set of 200 training-side problems, disjoint from the benchmark test sets, with $\delta_{\mathrm{dep}}=0.03$ (three percentage points). The reopening window spans steps 6,000--8,000 of autoregressive training. The ramp begins at the first check in this window satisfying the criterion, with step 8,000 as the latest scheduled start. Over the next 3,000 steps, $\alpha$ increases linearly from 0 to 1 and $w$ decreases linearly from 1 to $w_{\min}=0.5$.

Table~\ref{tab:reader-schedule} collects the complete internal-reader schedule. During parallel training, $p_r$ takes values $1.0$, $0.5$, and $0.3$ in successive epochs. The gate increases linearly from 0 to 1 during the second epoch as $w$ decreases linearly from 1 to 0.5; both remain at their endpoint values in the third epoch. At the start of autoregressive refinement, $\alpha=0$ and $w=1$, with $p_r=0.3$ throughout that phase. Refinement uses one block for its first 1,000 steps, then four blocks of eight latents, with producer masking $p_q=0.5$ after the first block. Decoupled training uses $w=1$ throughout and has no internal-reader gradient path. Epochs and steps are local to each phase.

\begin{table}[!htbp]
\centering
\caption{Unified-reader schedules. Epochs and steps are local to each phase.}
\label{tab:reader-schedule}
\small
\setlength{\tabcolsep}{5pt}
\renewcommand{\arraystretch}{1.1}
\begin{tabular*}{\linewidth}{@{\extracolsep{\fill}}lcc@{}}
\toprule
Setting & Parallel & Autoregressive\\
\midrule
Reader question mask $p_r$ & $1.0\,/\,0.5\,/\,0.3$ by epoch & 0.3\\
Gradient gate $\alpha$ & $0\rightarrow1$ in epoch 2 & $0\rightarrow1$ over 3,000 steps\\
External weight $w$ & $1\rightarrow0.5$ in epoch 2 & $1\rightarrow0.5$ over 3,000 steps\\
Readiness checks & -- & Every 500 steps; 200 problems\\
Reopening window & -- & Steps 6,000--8,000\\
\bottomrule
\end{tabular*}
\end{table}
\FloatBarrier

\paragraph{Inference.}
At inference, both question masks are disabled, $p_q=p_r=0$, and the producer generates $z^{\mathrm{ar}}$ from $x$ using its own preceding latent blocks. Decoupled CALR predicts from $p_{\bar\rho}(\cdot\mid P(z^{\mathrm{ar}}),\tau_{\mathrm{ans}})$; unified CALR predicts from $p_R(\cdot\mid x,P(z^{\mathrm{ar}}),\tau_{\mathrm{ans}})$ through its internal adapter in the same VLM. Answers are decoded greedily and compared with reference final answers. Rendering, the optical encoder, and reference derivations and latents serve training and representation evaluation; question-to-answer inference uses the question-conditioned producer and the chosen readout. $D_{\mathrm{ocr}}$ serves training and derivation evaluation, while $\bar\rho$ participates in answer prediction only in the decoupled realization.

\subsection{Training Data}
\label{app:training-data}

Training separates learning an optical representation of a derivation from learning to generate that representation from a question. The resampler uses the rendered derivations in Table~\ref{tab:training-data}; the external reader and both producer phases use GSM8K-Aug question--derivation--answer triples. MATH~\citep{hendrycks2021math} and MathX~\citep{rawal2024mathx} contribute only to reference-latent construction. Training examples are disjoint from the benchmark evaluation examples, including MATH-500.

\begin{table}[!htbp]
\centering
\caption{Training data and their roles in CALR.}
\label{tab:training-data}
\small
\setlength{\tabcolsep}{6pt}
\begin{tabular*}{\linewidth}{@{\extracolsep{\fill}}llr@{}}
\toprule
Source & Training use & Examples\\
\midrule
GSM8K-Aug & Resampler, reader, and producer & 384,618\\
MATH & Resampler & 11K\\
MathX-5M subset & Resampler & 818K\\
\bottomrule
\end{tabular*}
\end{table}

Resampler inputs contain the derivation text, with the separate question and answer fields omitted. GSM8K-Aug derivations are rendered in their original and natural-language forms, yielding approximately 1.6M encoded pages across the combined corpus. A held-out set of 2,701 chains supports transcription validation. The resampler ablation in Table~\ref{tab:resampler-ablation} is evaluated separately on the 1,319 GSM8K-Aug benchmark examples. Rendering and feature preparation are specified in Appendix~\ref{app:resampling-v13}.

\subsection{Evaluation Data}
\label{app:evaluation-data}

We evaluate on five mathematical reasoning benchmarks, from grade-school arithmetic to competition-level mathematics. Evaluation is zero-shot: the model receives the question without a reference derivation or worked examples. The reader and producer are trained on GSM8K-Aug; the remaining benchmarks evaluate transfer beyond this training distribution.

\begin{itemize}
\setlength{\itemsep}{4pt}
\item \textbf{GSM8K-Aug.} Evaluation uses the 1,319 GSM8K test problems~\citep{cobbe2021gsm8k}. These grade-school word problems require multi-step arithmetic and provide the in-distribution evaluation for the reader and producer. The same test set supports the diagnostic and ablation studies, with sampling specified in Appendix~\ref{app:analysis-aggregation}.
\item \textbf{GSM-Hard.} This benchmark~\citep{gao2023pal} contains 1,319 problems constructed by replacing numbers in the GSM8K test set with substantially larger quantities. It preserves the underlying problem structure while testing arithmetic reasoning under a numerical distribution shift.
\item \textbf{SVAMP.} Elementary arithmetic word problems are constructed through structural and linguistic variations of existing problems~\citep{patel2021svamp}. The benchmark tests whether a model follows the mathematical relationships expressed in the question as its wording and structure change.
\item \textbf{MultiArith.} This benchmark~\citep{roy2015multiarith} consists of word problems requiring multiple arithmetic operations. It evaluates the composition of elementary operations into a coherent solution beyond the GSM8K-Aug training distribution.
\item \textbf{MATH-500.} This 500-problem subset of the MATH test set~\citep{lightman2024verify} evaluates competition-level mathematical reasoning. Its evaluation examples are excluded from reference-latent training, and the reader and producer receive no MATH training problems. It therefore tests transfer from grade-school problem solving to more demanding mathematics.
\end{itemize}

\subsection{Base Models and Training Configurations}
\label{app:training-config}

\paragraph{Model roles and optimization.}
The base model $F$ supplies the producer and readout routes. Our principal configuration uses Qwen3-VL-4B-Instruct~\citep{bai2025qwen3vl}, with Qwen3-VL-2B-Instruct, Qwen3-VL-8B-Instruct, and LLaVA-V1.6-Mistral-7B~\citep{liu2024llavanext} used in the cross-model evaluations. The complete results also include the Qwen3-4B language model~\citep{yang2025qwen3} (Appendix~\ref{app:extended-results}). DeepSeek-OCR2 supplies the fixed optical encoder $E$ and decoder $D_{\mathrm{ocr}}$ used to construct and supervise the latent representation. Its optical feature dimension remains fixed, while the interface $P$ maps into the input dimension of the selected base model.

Training uses PyTorch~\citep{paszke2019pytorch}, Transformers~\citep{wolf2020transformers}, and PEFT~\citep{mangrulkar2022peft} on eight NVIDIA H200 GPUs (141\,GB each), with data parallelism and bfloat16 precision. All stages use AdamW~\citep{loshchilov2019adamw}, a peak learning rate of $10^{-4}$, cosine decay after linear warm-up, and gradient clipping at $1.0$. Optical page features are precomputed with the frozen encoder. Table~\ref{tab:training-config} gives the stage-specific settings. The LoRA~\citep{hu2022lora} adapters have rank 32 for the reader and rank 16 for the producer. In unified CALR, the internal reader inherits the Stage-I reader adapter, while its producer has the same configuration as in decoupled CALR.

\begin{table}[!htb]
\centering
\caption{Per-stage training configurations. Rows marked ``unified'' apply only to unified CALR; the decoupled form uses the same shared settings with $\alpha=0$, $w=1$, and no internal reader.}
\label{tab:training-config}
\begingroup
\fontsize{8}{9.5}\selectfont
\setlength{\tabcolsep}{2pt}
\renewcommand{\arraystretch}{1.16}
\begin{tabular*}{\linewidth}{@{\extracolsep{\fill}}>{\raggedright\arraybackslash}p{0.25\linewidth}*{4}{>{\centering\arraybackslash}p{0.174\linewidth}}@{}}
\toprule
Hyperparameter & Stage I:\newline Representation & Stage I:\newline Readout & Stage II:\newline Drafting & Stage II:\newline Refinement\\
\midrule
Training data & Rendered chains\newline (Appendix~\ref{app:training-data}) & GSM8K-Aug\newline 385K & GSM8K-Aug\newline 385K & GSM8K-Aug\newline 385K\\
Optimizer / LR schedule & AdamW / cosine & AdamW / cosine & AdamW / cosine & AdamW / cosine\\
Peak learning rate & $10^{-4}$ & $10^{-4}$ & $10^{-4}$ & $10^{-4}$\\
Warm-up steps & 300 & 200 & 360 & 360\\
Steps (epochs) & 12,000 ($\approx1$) & 6,000 (2) & 18,000 (3) & 18,000 (3)\\
Global batch size & 128 & 128 & 64 & 64\\
Adapter rank & -- & 32 & 16 & 16\\
Latents $K$ / blocks $B$ & 32 & 32 & 32 / 1 & 32 / 4\\
Block-wise masking $p_q$ & -- & -- & -- & 0.5 ($b>1$)\\
OCR weight $\lambda$ & -- & -- & $0.25\rightarrow0.05$ & $0.25{\to}0.05{\to}0.01$\\
Reader derivation weight $\eta$ & -- & 1 & $0.5\rightarrow0.1$ & 1\\
Resampler segment weight $\beta$
& 0.5 & -- & -- & --\\
AR block weight $\beta_{\mathrm{ref}}$
& -- & -- & -- & 1\\
Reader mask $p_r$ (unified) & -- & -- & $1.0\,/\,0.5\,/\,0.3$\newline by epoch & 0.3\\
Gradient gate $\alpha$ (unified) & -- & -- & $0\rightarrow1$\newline (epoch 2) & $0\rightarrow1$\newline (after admission)\\
External weight $w$ (unified) & -- & -- & $1\rightarrow0.5$\newline (epoch 2) & $1\rightarrow0.5$\newline (after admission)\\
Trainable modules & Resampler $\psi$ & Reader adapter,\newline mode tokens & $W_\theta$, $r_{1:K}$, tokens\newline ($+R$ if unified) & Same as drafting\\
Frozen modules & $E$, $D_{\mathrm{ocr}}$ & $P$, base VLM & $E$, $D_{\mathrm{ocr}}$, $\bar\rho$,\newline $P$, $H$, base VLM & Same as drafting\\
\bottomrule
\end{tabular*}
\par\vspace{3pt}
\begin{minipage}{\linewidth}
\fontsize{8}{9.5}\selectfont
385K denotes 384,618 examples. Token entries refer to trainable special-token embeddings; $R$ has rank 32. Refinement uses one block for its first 1,000 steps before switching to four. Table~\ref{tab:reader-schedule} specifies the admission timing and ramp.
\end{minipage}
\endgroup
\end{table}

\FloatBarrier
\paragraph{Functional anchoring.}
The resampler is trained for approximately one pass over the rendered corpus with $\beta=0.5$, $\gamma=0.01$, and target norm $\nu_0=6.5$ in Equation~\ref{eq:anchor-objective-v13}. Relative Gaussian noise of 5\% is applied to latent states during this stage. We then fix the resampler and train the external reader for two epochs with $\eta=1$. The interface scale $s$ is calibrated to match optical-latent and token-embedding norms; $U$, $P$, and $H$ remain fixed during reader and producer training. The trained reader is frozen as $\bar\rho$ for Stage II.

\paragraph{Parallel-to-autoregressive curriculum.}
Each producer phase runs for approximately three epochs over GSM8K-Aug. In parallel training, $\lambda$ decreases linearly from 0.25 to 0.05 during the first epoch and then remains fixed. The external-reader derivation weight $\eta$ is 0.5 in the first epoch, decreases to 0.1 during the second, and remains at 0.1 in the third. Autoregressive training warm-starts from the parallel-trained producer. It uses one block for the first 1,000 steps, followed by four blocks of eight latents; question masking with $p_q=0.5$ applies after the first block. Its OCR weight follows the three-epoch schedule $0.25\rightarrow0.05\rightarrow0.01$, with $\eta=1$ for the external reader. Equation~\ref{eq:autoregressive-objective-v13} combines full-trajectory supervision from $D_{\mathrm{ocr}}$ with block-level derivation recovery by the frozen external reader. We set $\beta_{\mathrm{ref}}=1$, so the block-level term follows the same $\lambda$ schedule as the full-trajectory OCR term; the reader's full-trajectory derivation loss retains $\eta=1$. Gradients remain connected across the generated prefix.

\subsection{General Evaluation Settings}
\label{app:general-evaluation}

Pass@1 and Acc. denote final-answer accuracy in percent. Reasoning length \#L counts intermediate text tokens or latent states, excluding the question and final answer. CALR uses 32 such states, while direct-answer methods have zero intermediate length. Average columns are unweighted means over the listed benchmarks; the Pass@1/\#L column divides the displayed mean accuracy by the displayed mean reasoning length, with a dash when the latter is zero. Character similarity and decoder cross-entropy evaluate reconstruction of a reference derivation, separately from answer accuracy (Appendix~\ref{app:analysis-decoding}).

\FloatBarrier
\section{Implementation Details of CALR}
\label{app:anchoring-v13}

\subsection{Rendering and Information-Balanced Resampling}
\label{app:resampling-v13}

Reference derivations are rendered in a monospaced font, with font sizes sampled from a Gaussian distribution truncated to 15--20 points. Text wraps at 70 characters on a 1,024-pixel-wide page whose height adapts to its content. The frozen DeepSeek-OCR2 encoder produces $N=256$ spatial features of dimension $d=1280$; its decoder also remains fixed throughout anchoring.

Index $V_1,\ldots,V_N$ in the encoder's spatial sequence order. Let $A(t)=\sum_{i=1}^{t}a_i$ be the cumulative feature-surprisal score, with $A(0)=0$. The segmentation in Equation~\ref{eq:anchored-construction-v13} takes the form
\begin{equation}
0=t_0<t_1<\cdots<t_K=N,\qquad
S_g=\{t_{g-1}+1,\ldots,t_g\},\qquad
A(t_g)\approx\frac{g}{K}A(N).
\label{eq:segment-boundaries-v13}
\end{equation}
The feature statistics remain fixed across training, while boundaries are recomputed for each rendered derivation. The score is the nonconstant quadratic component of diagonal-Gaussian negative log likelihood. Equal cumulative score allocates fewer spatial positions to information-dense regions and more positions to sparse regions.

The resampler contextualizes the complete feature sequence through four self-attention layers. A shared query then cross-attends to the contextualized features within each $S_g$, using an intra-segment position bias. Each output therefore represents its designated region with access to global page context.

\subsection{Supervision Targets and Canvas Construction}
\label{app:canvas-v13}

Spatial segmentation and text-target construction are separate operations. Spatial segments follow the feature-surprisal scores above; text spans are formed from paragraph and punctuation boundaries in the reference derivation. These spans retain their reading order when assigned as supervision for latent segments. We write $c_g$ for the target associated with latent position $g$. For a block of positions $\mathcal B_b$, its target is $c_{\mathcal B_b}=\operatorname{concat}_{g\in\mathcal B_b}c_g$.

The optical latent dimension matches the decoder's visual feature dimension, $d_z=d$. Let $V^{\mathrm{blank}}\in\mathbb R^{N\times d_z}$ be the features of a rendered blank page, and let $\pi:\{1,\ldots,K\}\rightarrow\{1,\ldots,N\}$ denote the fixed injective placement of latent positions on the decoder's visual grid. For a subset $J$ of latent positions, define
\begin{equation}
\left[\operatorname{Canvas}_J(z_J)\right]_i=
\begin{cases}
z_g, & i=\pi(g),\quad g\in J,\\
V_i^{\mathrm{blank}}, & i\notin\pi(J).
\end{cases}
\label{eq:canvas-placement-v13}
\end{equation}
The complete, single-latent, and block canvases take $J=\{1,\ldots,K\}$, $J=\{g\}$, and $J=\mathcal B_b$, respectively; we write $\operatorname{Canvas}(z)$ for the complete canvas. Partial canvases retain absolute positions under the same placement map. Unused positions contain blank-page features rather than zero vectors.

During Stage~I, the frozen DeepSeek-OCR2 decoder supervises both the complete reference canvas and single-latent canvases, together with norm regularization. During Stage~II, full-trajectory OCR supervision is retained in both producer phases, while autoregressive refinement additionally uses the frozen external reader to recover aligned derivation spans from block canvases. The reader's full-trajectory answer and derivation losses remain active in both phases. The optical readout $\widehat c(z)$ is used to evaluate C2, while $\mathrm{Read}_{\mathrm{out}}$ provides additional block-level derivation supervision during autoregressive training.

\subsection{Fixed Interfaces}
\label{app:interfaces-v13}

The maps $P:\mathbb R^{d_z}\rightarrow\mathbb R^{d_{\mathrm{emb}}}$ and $H:\mathbb R^{d_{\mathrm{emb}}}\rightarrow\mathbb R^{d_z}$ act position-wise. We construct $U$ by applying a reduced QR decomposition to a randomly initialized matrix $M\in\mathbb R^{d_{\mathrm{emb}}\times d_z}$:
\[
M=UT,\qquad U^\top U=I_{d_z},
\]
where $T\in\mathbb R^{d_z\times d_z}$ is upper triangular. We retain the orthonormal factor $U$ and fix it throughout reader and producer training. With a positive calibration scale $s$, the interfaces satisfy
\begin{equation}
\begin{aligned}
P&=sU,\qquad H=s^{-1}U^\top,\\
HP&=I_{d_z},\qquad PH=UU^\top.
\end{aligned}
\label{eq:interface-dimensions-v13}
\end{equation}
Thus, $H$ recovers the original coordinates of an embedded optical latent, while $PH$ projects a VLM state onto the interface subspace. In both producer phases, $H$ converts selected query states into optical latents. The OCR canvas receives $z$ directly, both readout routes and the producer feedback route receive $P(z)$. The interfaces remain fixed during training. For block-canvas supervision, $\mathrm{Read}_{\mathrm{out}}$ includes the same fixed reader-input interface and the derivation mode $\tau_{\mathrm{cot}}$.

\section{Diagnostic and Ablation Protocols}
\label{app:latent-analysis}
\label{app:prelim-analysis-v13}

\subsection{Sampling and Answer-Input Interventions}
\label{app:analysis-aggregation}

Figure~\ref{fig:prelim-interventions-v13} and Tables~\ref{tab:analysis-interventions}--\ref{tab:analysis-attention} use the same fixed subset of 300 problems sampled at random from the 1,319 GSM8K-Aug evaluation problems. RoT uses the released model, DLR-SFT uses our reproduction, and CALR uses the unified realization. CALR percentages average repeated evaluations on this subset, including sampled replacements for stochastic interventions. The remaining diagnostic and ablation evaluations use the complete GSM8K-Aug evaluation set.

For a dataset $\mathcal D$, let $A_{\mathrm{base}}$ denote accuracy with each problem's native latents. An intervention $\iota$ replaces $G_\theta(x)$ by a possibly stochastic state $z^\iota\sim\iota(G_\theta(x))$ while keeping the answer model fixed. We define
\begin{equation}
\begin{aligned}
A_{\mathrm{base}}
&=\Pr_{(x,y)\sim\mathcal D}
\left[\arg\max_{y'}p_\phi(y'\mid x,G_\theta(x))=y\right],\\
A_\iota
&=\mathbb E_{(x,y)\sim\mathcal D}
\mathbb E_{z^\iota\sim\iota(G_\theta(x))}
\mathbf 1\!\left[\arg\max_{y'}p_\phi(y'\mid x,z^\iota)=y\right],\\
\Lambda_\iota(z\rightarrow\hat y)
&=A_{\mathrm{base}}-A_\iota.
\end{aligned}
\label{eq:intervention-v13}
\end{equation}
We apply two intervention families. \emph{Content interventions} preserve the latent interface but replace its native trajectory: $I_{\mathrm{swap}}$ supplies another problem's latents, and $I_{\mathrm{const}}$ supplies one fixed latent sequence to every problem. \emph{Interface interventions} disrupt the channel itself: $I_{\mathrm{rand}}$ supplies norm-matched random states, and $I_{\mathrm{del}}$ removes the latent slots. Writing $\Lambda_{\mathrm{swap}}:=\Lambda_{I_{\mathrm{swap}}}$ and analogously for the remaining interventions, their aggregate effects are
\begin{equation}
\begin{gathered}
\mathcal I_{\mathrm{content}}=\{I_{\mathrm{swap}},I_{\mathrm{const}}\},
\qquad
\mathcal I_{\mathrm{interface}}=\{I_{\mathrm{rand}},I_{\mathrm{del}}\},\\
\Lambda_{\mathrm{content}}
=\tfrac12(\Lambda_{\mathrm{swap}}+\Lambda_{\mathrm{const}}),
\qquad
\Lambda_{\mathrm{interface}}
=\tfrac12(\Lambda_{\mathrm{rand}}+\Lambda_{\mathrm{del}}).
\end{gathered}
\label{eq:families-v13}
\end{equation}
The deletion effect $\Lambda_{\mathrm{del}}$ directly operationalizes C1 at the accuracy level, while $\Lambda_{\mathrm{interface}}$ summarizes the model's broader sensitivity to disruption of the latent input. Content interventions test whether an active interface carries problem-specific information, while decoded-derivation validity evaluates C2. As a control on the competing direct route, we withhold $x$ from the answer model while retaining the latents generated from the original question,
\begin{equation}
A_{-x}
=\Pr_{(x,y)\sim\mathcal D}
\left[\arg\max_{y'}p_\phi(y'\mid\varnothing_x,G_\theta(x))=y\right],
\qquad
\Lambda_x=A_{\mathrm{base}}-A_{-x}.
\label{eq:question-effect-v13}
\end{equation}

Each native trajectory is generated from its own question before modifying the answer model's inputs. DLR-SFT uses six pooled latent codes in these input interventions.

When accuracies are expressed as probabilities, intervention effects are reported in percentage points as $100\Lambda$. The unchanged-answer rate compares each swapped prediction with the corresponding native prediction. For an evaluation with $n$ examples,
\begin{equation}
U_{\mathrm{swap}}=\frac{100}{n}\sum_{i=1}^{n}
\mathbf 1[\widehat y_i^{\mathrm{swap}}=\widehat y_i^{\mathrm{base}}].
\label{eq:answer-consistency}
\end{equation}
The reported rate averages these comparisons over evaluations, independently of whether the shared answer is correct.

\subsection{Attention and Gradient Measurements}
\label{app:analysis-probes}

Let $a_z^{(i)}$ and $a_x^{(i)}$ denote the answer-token attention masses assigned to latent and question positions for example $i$. Table~\ref{tab:analysis-attention} reports the mean of the per-example normalized latent shares,
\begin{equation}
\overline a_{\mathrm{latent}}
=\frac{100}{n}\sum_{i=1}^{n}
\frac{a_z^{(i)}}{a_z^{(i)}+a_x^{(i)}}.
\label{eq:normalized-latent-attention}
\end{equation}
Normalization includes the question and latent groups and precedes averaging over examples. Figure~\ref{fig:prelim-interventions-v13} instead presents group means without renormalizing over question and latent positions; Table~\ref{tab:prelim-measurements-v13} collects these values alongside the gradient measurements.

For the answer loss $\ell_{\mathrm{ans}}=-\log p_\phi(y\mid x,z)$, let $e_x$ denote the question-input embeddings. We compute
\begin{equation}
\overline g=\frac{1}{n}\sum_{i=1}^{n}
\frac{\lVert\nabla_z\ell_{\mathrm{ans}}^{(i)}\rVert_2}
{\lVert\nabla_{e_x}\ell_{\mathrm{ans}}^{(i)}\rVert_2}.
\label{eq:analysis-gradient-ratio}
\end{equation}
For discrete latents, differentiation is with respect to their continuous input embeddings. Each norm includes all positions and feature dimensions in its input group.

\begin{table}[!htbp]
\centering
\caption{Answer-stage attention and gradient measurements.}
\label{tab:prelim-measurements-v13}
\small
\setlength{\tabcolsep}{5pt}
\begin{tabular}{lccc}
\toprule
System & Question attention & Latent attention & Latent/question gradient ratio\\
\midrule
RoT & $0.51$ & $0.33$ & $1.12$\\
DLR-SFT & $0.72$ & $0.05$ & $0.13$\\
\bottomrule
\end{tabular}
\end{table}

\subsection{Decoded Chains and arithmetic-step recovery}
\label{app:analysis-decoding}

All systems in Table~\ref{tab:analysis-reasoning} are decoded with the frozen DeepSeek-OCR2 decoder, which receives their optical latent representations without question text. Let $\widehat c_i$ and $c_i$ denote the decoded and reference chains. Character similarity uses Python's \href{https://docs.python.org/3/library/difflib.html\#difflib.SequenceMatcher.ratio}{\texttt{difflib.SequenceMatcher.ratio()}}, which recursively matches contiguous character blocks. If $M(\widehat c_i,c_i)$ counts the matched characters, the score is
\begin{equation}
\operatorname{Sim}(\widehat c_i,c_i)
=\frac{2M(\widehat c_i,c_i)}{|\widehat c_i|+|c_i|}.
\label{eq:chain-character-similarity}
\end{equation}
Lengths are character counts, and scores are averaged over problems.

For arithmetic-step recovery, we use each reference result in $c_i$ to locate candidate passages in $\widehat c_i$, then verify the corresponding operands and arithmetic relation within the local clause or sentence. Adjacent context and previously matched results help resolve explicit references. Matching normalizes numeric formatting and arithmetic symbols and allows natural-language expressions, reversed equalities, and commutative operand order. Thus, $4+24=28$ and ``adding 4 to 24 gives 28'' match the same step, whereas merely mentioning $28$ does not. Let $m_i$ denote the number of arithmetic steps in the reference chain $c_i$, and let $\mathcal{A}_i$ be a maximum-cardinality one-to-one matching between the arithmetic steps in $c_i$ and those expressed in $\widehat c_i$ under these rules. We compute
\begin{equation}
\operatorname{Rec}(\widehat c_i,c_i)
=
\frac{|\mathcal{A}_i|}{m_i}.
\label{eq:intermediate-value-recovery}
\end{equation}
Scores are averaged across problems. The \emph{same-problem} evaluation compares $\widehat c_i$ with $c_i$, whereas the \emph{other-problem} evaluation holds $\widehat c_i$ fixed and applies the same procedure to another problem's reference $c_{j(i)}$, where $j(i)\neq i$. Answer accuracy is evaluated under the common answer-evaluation setting used by the three systems and reported separately from chain-content metrics.

\subsection{Generation-Feedback Evaluation}
\label{app:analysis-feedback}

Final diagnostics use the paired feedback construction and boundary-averaged effect in Appendix~\ref{app:calr-realizations}, with all 1,319 GSM8K-Aug problems and question masks disabled. Each system is evaluated through its prescribed answer route, using the same readout within each native--intervened pair.

Table~\ref{tab:analysis-feedback} compares the autoregressively trained producer with the parallel-trained control under the same four-block rollout, with $K=32$ latents in four blocks of eight. Zero replaces feedback embeddings with zeros. Shuffle implements $\mathrm{swap}$ by a derangement across the example dimension within a rollout microbatch of 16: each problem receives another problem's prefix while latent-position order is preserved. Question tokens and latent queries remain unchanged within each pair, and all subsequent blocks are generated freely.

The all-latent view reads the complete 32-latent trajectory; the last-block view supplies only the final eight generated latents. Each effect is measured relative to the native accuracy for that view and averaged over intervention boundaries, as in Equation~\ref{eq:autoregressive-effect-v13}. The last-block comparison reads $\mathbf z_B$ and $\widehat{\mathbf z}_{B\mid b}^{\mathcal S}$ under the same readout protocol.

\subsection{Frozen-Reader Evaluation}
\label{app:analysis-reader}

Table~\ref{tab:analysis-reader} fixes the external readout $\bar\rho$ and withholds the question. Its four inputs are the raw DeepSeek-OCR2 features $V$ of the rendered reference derivation, their resampled reference trajectory $z^\star$, the problem-conditioned output $G_\theta(x)$ of the unified-trained CALR producer, and blank-page features $V^{\mathrm{blank}}$. Each optical feature sequence passes through the same fixed interface $P$. The 62.5\% accuracy therefore measures the unified-trained producer's performance through the external reader $\bar\rho$.

\subsection{Component Ablations}
\label{app:component-ablations}

\paragraph{Training components.}
Table~\ref{tab:component-ablation} evaluates unified CALR on GSM8K-Aug with 32 latents. Every variant retains the internal reader. External supervision uses the frozen question-blind readout, and the reader-mask column specifies whether $p_r$ is active in the current training phase. Autoregressive variants warm-start from externally supervised parallel training. A dash in the AR-supervision column indicates that the configuration contains only the parallel phase.

\paragraph{Resampling strategies.}
Table~\ref{tab:resampler-ablation} compares three ways to resample optical features from rendered reference derivations. Global cross-attention operates on the complete feature sequence; fixed anchors assign equal feature counts to each output latent; dynamic anchors adjust segment widths to balance cumulative information. The trained representations are decoded by DeepSeek-OCR2 on GSM8K-Aug. Similarity uses Equation~\ref{eq:chain-character-similarity}, reported in percent; CE is decoder cross-entropy against the reference derivation.

\paragraph{Output and feedback interfaces.}
In Table~\ref{tab:analysis-interface}, CALR uses the fixed maps $P=sU$ and $H=s^{-1}U^\top$, with $U^\top U=I_{d_z}$ as in Equation~\ref{eq:fixed-interface-v13}. The second configuration replaces $H$ with a trainable MLP and uses a frozen copy of the reader projection for feedback. The third trains both the output head and the feedback projection. The producer parameters $W_\theta$ remain trainable in all three configurations. Each uses four-block autoregressive rollouts and the paired evaluation in Appendix~\ref{app:analysis-feedback}, with interventions applied to the actual feedback embeddings after projection. Accuracies and prefix effects use the same readout protocol across interface configurations.
\Needspace{.8\textheight}
\section{Further Experiments and Visualization Analysis}
\label{app:further-experiments}

\subsection{Complete Benchmark Results}
\label{app:extended-results}

Table~\ref{tab:main-results-full} extends Table~\ref{tab:main-results-v14} with the Qwen3-4B language model, iCoT, and DLR's reinforcement-learning variant. All CALR configurations use supervised training.

\setlength{\intextsep}{6pt}
\begin{table}[!htb]
\centering
\caption{Complete results on four grade-school reasoning benchmarks. Acc.\ denotes accuracy (\%), and \#L denotes reasoning length. Average is the unweighted mean across the four benchmarks.}
\label{tab:main-results-full}
\begingroup
\fontsize{8}{10}\selectfont
\setlength{\tabcolsep}{1.8pt}
\renewcommand{\arraystretch}{1.06}
\newcommand{\MRMethodLabel}{Method}
\newcommand{\MRAverageLabel}{Average}
\sbox{\MRtablebox}{%
\begin{tabular}{@{}l*{10}{c}@{}}
\toprule
& \multicolumn{2}{c}{\textbf{GSM8K-Aug}}
& \multicolumn{2}{c}{\textbf{GSM-Hard}}
& \multicolumn{2}{c}{\textbf{SVAMP}}
& \multicolumn{2}{c}{\textbf{MultiArith}}
& \multicolumn{2}{c}{\textbf{\MRAverageLabel}}\\
\cmidrule(lr){2-3}\cmidrule(lr){4-5}\cmidrule(lr){6-7}\cmidrule(lr){8-9}\cmidrule(l){10-11}
\textbf{\MRMethodLabel} & Acc. & \#L & Acc. & \#L & Acc. & \#L & Acc. & \#L & Acc. & \#L\\
\midrule
\rowcolor{MRgroup}
\multicolumn{11}{l}{\strut\bfseries\itshape Qwen3-4B (LLM)}\\
Direct-SFT & \MRcell{27.4}{.4}{} & \MRcell{0.0}{.0}{} & \MRcell{9.2}{.2}{} & \MRcell{0.0}{.0}{} & \MRcell{69.4}{.4}{} & \MRcell{0.0}{.0}{} & \MRcell{86.3}{.3}{} & \MRcell{0.0}{.0}{} & $48.1$ & $0.0$\\
CoT-SFT & \MRcell{83.1}{.3}{} & \MRcell{119.8}{.8}{} & \MRcell{52.3}{.3}{} & \MRcell{189.4}{1.1}{} & \MRcell{82.9}{.3}{} & \MRcell{53.7}{.7}{} & \MRcell{98.5}{.3}{} & \MRcell{57.6}{.8}{} & $79.2$ & $105.1$\\
\midrule
iCoT & \MRcell{13.5}{.2}{} & \MRcell{0.0}{.0}{} & \MRcell{4.1}{.2}{} & \MRcell{0.0}{.0}{} & \MRcell{36.9}{.2}{} & \MRcell{0.0}{.0}{} & \MRcell{49.2}{.7}{} & \MRcell{0.0}{.0}{} & $25.9$ & $0.0$\\
Coconut (CoLM 2025) & \MRcell{16.9}{.3}{} & \MRcell{6.0}{.0}{} & \MRcell{5.4}{.3}{} & \MRcell{6.0}{.0}{} & \MRcell{43.6}{.5}{} & \MRcell{6.0}{.0}{} & \MRcell{60.3}{.7}{} & \MRcell{6.0}{.0}{} & $31.6$ & $6.0$\\
CODI (EMNLP 2025) & \MRcell{7.3}{.5}{} & \MRcell{6.0}{.0}{} & \MRcell{2.2}{.2}{} & \MRcell{6.0}{.0}{} & \MRcell{11.0}{.6}{} & \MRcell{6.0}{.0}{} & \MRcell{18.3}{.8}{} & \MRcell{6.0}{.0}{} & $9.7$ & $6.0$\\
CoLaR (NeurIPS 2025) & \MRcell{40.0}{.2}{} & \MRcell{39.6}{.1}{} & \MRcell{9.2}{.1}{} & \MRcell{47.4}{.2}{} & \MRcell{57.7}{.2}{} & \MRcell{19.2}{.1}{} & \MRcell{82.2}{.1}{} & \MRcell{21.1}{.1}{} & $47.3$ & $31.8$\\
DLR (SFT) & \MRcell{47.4}{.4}{} & \MRcell{6.4}{.3}{} & \MRcell{13.7}{.3}{} & \MRcell{6.8}{.2}{} & \MRcell{70.3}{.2}{} & \MRcell{3.6}{.3}{} & \MRcell{96.1}{.3}{} & \MRcell{4.2}{.1}{} & $56.9$ & $5.3$\\
DLR (RL) & \MRcell{62.2}{.4}{*} & \MRcell{6.5}{.2}{} & \MRcell{\MRbest{19.3}}{.3}{*} & \MRcell{7.0}{.3}{} & \MRcell{72.3}{.2}{*} & \MRcell{3.8}{.2}{} & \MRcell{97.8}{.3}{*} & \MRcell{4.3}{.3}{} & $62.9$ & $5.4$\\
\midrule
\rowcolor{MRdecoupled}
\textit{Ours, decoupled (Par.)} & \MRcell{49.8}{.3}{} & \MRcell{32.0}{.0}{} & \MRcell{14.5}{.4}{} & \MRcell{32.0}{.0}{} & \MRcell{70.1}{.2}{} & \MRcell{32.0}{.0}{} & \MRcell{96.8}{.2}{} & \MRcell{32.0}{.0}{} & $57.8$ & $32.0$\\
\rowcolor{MRunified}
\textit{Ours, unified (Par.)} & \MRcell{51.2}{.3}{} & \MRcell{32.0}{.0}{} & \MRcell{15.1}{.3}{} & \MRcell{32.0}{.0}{} & \MRcell{71.5}{.2}{} & \MRcell{32.0}{.0}{} & \MRcell{97.4}{.1}{} & \MRcell{32.0}{.0}{} & $58.8$ & $32.0$\\
\rowcolor{MRdecoupled}
\textit{Ours, decoupled (AR)} & \MRcell{60.7}{.2}{} & \MRcell{32.0}{.0}{} & \MRcell{16.9}{.3}{} & \MRcell{32.0}{.0}{} & \MRcell{74.8}{.3}{} & \MRcell{32.0}{.0}{} & \MRcell{98.0}{.2}{} & \MRcell{32.0}{.0}{} & $62.6$ & $32.0$\\
\rowcolor{MRunified}
\textit{Ours, unified (AR)} & \MRcell{\MRbest{62.3}}{.3}{} & \MRcell{32.0}{.0}{} & \MRcell{17.6}{.3}{} & \MRcell{32.0}{.0}{} & \MRcell{\MRbest{76.2}}{.2}{} & \MRcell{32.0}{.0}{} & \MRcell{\MRbest{98.8}}{.1}{} & \MRcell{32.0}{.0}{} & $\MRbest{63.7}$ & $32.0$\\
\midrule
\rowcolor{MRgroup}
\multicolumn{11}{l}{\strut\bfseries\itshape Qwen3-VL-4B-Instruct (LMM)}\\
Direct-SFT & \MRcell{26.2}{.3}{} & \MRcell{0.0}{.0}{} & \MRcell{9.5}{.1}{} & \MRcell{0.0}{.0}{} & \MRcell{70.0}{.3}{} & \MRcell{0.0}{.0}{} & \MRcell{85.6}{.4}{} & \MRcell{0.0}{.0}{} & $47.8$ & $0.0$\\
CoT-SFT & \MRcell{81.2}{.4}{} & \MRcell{127.3}{1.0}{} & \MRcell{53.4}{.3}{} & \MRcell{191.1}{1.6}{} & \MRcell{84.3}{.3}{} & \MRcell{55.9}{.6}{} & \MRcell{98.3}{.4}{} & \MRcell{59.1}{.6}{} & $79.3$ & $108.4$\\
\midrule
Coconut (CoLM 2025) & \MRcell{16.1}{.4}{} & \MRcell{6.0}{.0}{} & \MRcell{5.2}{.3}{} & \MRcell{6.0}{.0}{} & \MRcell{42.0}{.3}{} & \MRcell{6.0}{.0}{} & \MRcell{58.9}{.2}{} & \MRcell{6.0}{.0}{} & $30.6$ & $6.0$\\
CODI (EMNLP 2025) & \MRcell{7.1}{.3}{} & \MRcell{6.0}{.0}{} & \MRcell{2.0}{.4}{} & \MRcell{6.0}{.0}{} & \MRcell{10.7}{.1}{} & \MRcell{6.0}{.0}{} & \MRcell{17.8}{.2}{} & \MRcell{6.0}{.0}{} & $9.4$ & $6.0$\\
CoLaR (NeurIPS 2025) & \MRcell{39.2}{.2}{} & \MRcell{38.8}{.5}{} & \MRcell{8.2}{.4}{} & \MRcell{47.8}{.7}{} & \MRcell{52.0}{.4}{} & \MRcell{18.4}{.6}{} & \MRcell{81.0}{.3}{} & \MRcell{22.1}{.3}{} & $45.1$ & $31.8$\\
RoT (ACL 2026) & \MRcell{37.8}{.3}{} & \MRcell{32.0}{.0}{} & \MRcell{14.1}{.2}{} & \MRcell{32.0}{.0}{} & \MRcell{72.7}{.4}{} & \MRcell{32.0}{.0}{} & \MRcell{97.2}{.4}{} & \MRcell{32.0}{.0}{} & $55.5$ & $32.0$\\
DLR (SFT) & \MRcell{48.0}{.4}{} & \MRcell{6.3}{.4}{} & \MRcell{14.7}{.3}{} & \MRcell{6.8}{.5}{} & \MRcell{68.7}{.2}{} & \MRcell{3.6}{.1}{} & \MRcell{97.8}{.3}{} & \MRcell{4.2}{.2}{} & $57.3$ & $5.2$\\
DLR (RL) & \MRcell{63.3}{.3}{*} & \MRcell{6.5}{.1}{} & \MRcell{\MRbest{18.3}}{.4}{*} & \MRcell{7.0}{.6}{} & \MRcell{72.7}{.3}{} & \MRcell{3.8}{.4}{} & \MRcell{98.3}{.2}{*} & \MRcell{4.3}{.3}{} & $63.2$ & $5.4$\\
\midrule
\rowcolor{MRdecoupled}
\textit{Ours, decoupled (Par.)} & \MRcell{51.2}{.2}{} & \MRcell{32.0}{.0}{} & \MRcell{13.8}{.3}{} & \MRcell{32.0}{.0}{} & \MRcell{71.3}{.2}{} & \MRcell{32.0}{.0}{} & \MRcell{97.6}{.1}{} & \MRcell{32.0}{.0}{} & $58.5$ & $32.0$\\
\rowcolor{MRunified}
\textit{Ours, unified (Par.)} & \MRcell{52.4}{.3}{} & \MRcell{32.0}{.0}{} & \MRcell{14.3}{.2}{} & \MRcell{32.0}{.0}{} & \MRcell{72.4}{.1}{} & \MRcell{32.0}{.0}{} & \MRcell{98.1}{.1}{} & \MRcell{32.0}{.0}{} & $59.3$ & $32.0$\\
\rowcolor{MRdecoupled}
\textit{Ours, decoupled (AR)} & \MRcell{62.1}{.2}{} & \MRcell{32.0}{.0}{} & \MRcell{16.1}{.2}{} & \MRcell{32.0}{.0}{} & \MRcell{76.6}{.3}{} & \MRcell{32.0}{.0}{} & \MRcell{98.8}{.1}{} & \MRcell{32.0}{.0}{} & $63.4$ & $32.0$\\
\rowcolor{MRunified}
\textit{Ours, unified (AR)} & \MRcell{\MRbest{63.8}}{.2}{} & \MRcell{32.0}{.0}{} & \MRcell{16.8}{.3}{} & \MRcell{32.0}{.0}{} & \MRcell{\MRbest{77.4}}{.1}{} & \MRcell{32.0}{.0}{} & \MRcell{\MRbest{99.7}}{.1}{} & \MRcell{32.0}{.0}{} & $\MRbest{64.4}$ & $32.0$\\
\bottomrule
\end{tabular}
\relax{}}
\ifdim\wd\MRtablebox>\linewidth
\resizebox{\linewidth}{!}{\usebox{\MRtablebox}}%
\else
\usebox{\MRtablebox}%
\fi
\par\vspace{4pt}
\begin{minipage}{\linewidth}
\fontsize{7.5}{9.2}\selectfont
\textbf{Bold} marks the highest accuracy among latent reasoning methods within each model group.
Blue and green shading distinguish decoupled and unified CALR.
Par.\ and AR denote parallel drafting and autoregressive refinement, respectively.
Asterisks retain the significance markers from DLR's original comparison.
\end{minipage}
\endgroup
\end{table}

\subsection{Analysis of Latent Reasoning}
\label{sec:latent-analysis}
\begingroup
\setlength{\intextsep}{7pt plus 1pt minus 1pt}
\setlength{\textfloatsep}{9pt plus 1pt minus 1pt}
\setlength{\floatsep}{8pt plus 1pt minus 1pt}
\definecolor{ANhighlight}{HTML}{E5EFFA}
\newsavebox{\ANleftcaption}
\newsavebox{\ANrightcaption}
\newlength{\ANcaptionheight}
\newcommand{\ANpair}[6]{%
  \sbox{\ANleftcaption}{\begin{minipage}[t]{#1}#3\end{minipage}}%
  \sbox{\ANrightcaption}{\begin{minipage}[t]{#2}#5\end{minipage}}%
  \setlength{\ANcaptionheight}{\dimexpr\ht\ANleftcaption+\dp\ANleftcaption\relax}%
  \ifdim\dimexpr\ht\ANrightcaption+\dp\ANrightcaption\relax>\ANcaptionheight
    \setlength{\ANcaptionheight}{\dimexpr\ht\ANrightcaption+\dp\ANrightcaption\relax}%
  \fi
  \begin{minipage}[t]{#1}\vspace{0pt}%
    \parbox[t][\ANcaptionheight][t]{\linewidth}{\usebox{\ANleftcaption}}\par
    \centering #4\par
  \end{minipage}\hfill
  \begin{minipage}[t]{#2}\vspace{0pt}%
    \parbox[t][\ANcaptionheight][t]{\linewidth}{\usebox{\ANrightcaption}}\par
    \centering #6\par
  \end{minipage}%
}
\newcommand{\ANformat}[1]{%
  \fontsize{9}{11}\selectfont
  \setlength{\tabcolsep}{#1}%
  \renewcommand{\arraystretch}{1.06}%
}
\newcommand{\ANInterventionRows}{%
RoT & $43.0$ & $42.0$ & $43.0$ & $1.0$ & $1.0$ & $2.0$ & $98.0$\\
DLR-SFT & $53.0$ & $53.0$ & $53.0$ & $52.0$ & $51.0$ & $1.0$ & $93.0$\\
\rowcolor{ANhighlight}
\textbf{CALR} & $\MRbest{64.3}$ & $1.4$ & $1.1$ & $1.7$ & $0.8$ & $\MRbest{61.9}$ & $\MRbest{1.3}$\\
}
\newcommand{\ANAttentionRows}{%
RoT & $39$ & $1.12$\\
DLR-SFT & $6$ & $0.13$\\
\rowcolor{ANhighlight}
\textbf{CALR} & $\MRbest{69}$ & $0.26$\\
}
\newcommand{\ANReasoningRows}{%
RoT & $0.07$ & $0.01$ & $0.01$ & $38.2$\\
DLR-SFT{*} & $0.09$ & $0.53$ & $0.13$ & $44.0$\\
\rowcolor{ANhighlight}
\textbf{CALR} & $\MRbest{0.34}$ & $\MRbest{0.74}$ & $0.07$ & $\MRbest{63.5}$\\
}
\newcommand{\ANReaderRows}{%
\ANRaw & $V$ & $98.9$\\
\ANReference & $z^\star$ & $96.7$\\
\rowcolor{ANhighlight}
\ANGenerated & $G_\theta(x)$ & $\MRbest{62.5}$\\
\ANBlank & $V^{\mathrm{blank}}$ & $0.5$\\
}
\newcommand{\ANFeedbackRows}{%
\rowcolor{ANhighlight}
\textbf{CALR} & \ANAll & $\MRbest{63.8}$ & $38.7$ & $45.4$ & $+25.1$ & $+18.4$\\
\rowcolor{ANhighlight}
 & \ANLast & $28.2$ & $6.1$ & $9.8$ & $+22.1$ & $+18.4$\\
\ANParallel & \ANAll & $51.1$ & $51.3$ & $51.0$ & $-0.2$ & $+0.1$\\
}
\newcommand{\ANInterfaceRows}{%
\rowcolor{ANhighlight}
\ANFixed & \ANFixedTied & $\MRbest{63.8}$ & $\MRbest{+25.1}$ & $\MRbest{+18.4}$\\
\ANMLP & \ANFrozenCopy & $59.2$ & $+15.2$ & $+13.3$\\
\ANTrainable & \ANTrainable & $56.33$ & $+6.2$ & $+3.5$\\
}
\newcommand{\ANfit}[1]{%
  \sbox{\MRtablebox}{#1}%
  \ifdim\wd\MRtablebox>\linewidth
    \resizebox{\linewidth}{!}{\usebox{\MRtablebox}}%
  \else
    \usebox{\MRtablebox}%
  \fi
}
We examine whether CALR makes latent reasoning both load-bearing for the answer and specific to the problem, then study how autoregressive training and fixed interfaces support this behavior. Accuracies are percentages and intervention effects are percentage-point drops. Appendix~\ref{app:latent-analysis} gives the measurement and evaluation details.

\paragraph{Latent reliance.}
Table~\ref{tab:analysis-interventions} extends the preliminary interventions to unified CALR. Its native accuracy is 64.3\%, but replacing or removing its latents reduces accuracy to 0.8--1.7\%, with the question still available. Under cross-problem replacement, only 1.3\% of its answers remain unchanged, compared with 98\% for RoT and 93\% for DLR-SFT. CALR therefore depends on the trajectory associated with the current problem, rather than merely the presence of latent slots. Removing the answer-side question instead preserves 61.9\% accuracy, a drop of only 2.4 points. Together, these interventions establish the load-bearing role of CALR's latents (C1) and show that they retain the information needed for answer prediction.

Table~\ref{tab:analysis-attention} characterizes the corresponding answer-stage interaction. CALR allocates 69\% of its attention over question and latent inputs to the latents, versus 39\% for RoT and 6\% for DLR-SFT. Its latent-to-question gradient ratio is 0.26, between DLR-SFT's 0.13 and RoT's 1.12. RoT thus has the largest relative latent gradient, while CALR has the strongest dependence on problem-specific latent content. The intervention and attention measurements together identify an answer process that actively uses the generated trajectory.

\begin{table}[!htbp]
\centering
\ANpair{.65\linewidth}{.32\linewidth}
{\caption{Answer-input interventions (\%). $U_{\mathrm{swap}}$: unchanged-answer rate under swap; No $x$: answer-side question removed.}\label{tab:analysis-interventions}}
{\begingroup\ANformat{2.2pt}
\ANfit{\begin{tabular}{l*{7}{c}}
\toprule
 & & \multicolumn{2}{c}{Content} & \multicolumn{2}{c}{Interface} & & \\
\cmidrule(lr){3-4}\cmidrule(lr){5-6}
System & Base & Swap & Const. & Rand. & Del. & No $x$ & $U_{\mathrm{swap}}$\\
\midrule
\ANInterventionRows
\bottomrule
\end{tabular}}
\endgroup}
{\caption{Answer-stage latent attention and latent-to-question gradient-norm ratios.}\label{tab:analysis-attention}}
{\begingroup\ANformat{1.6pt}
\ANfit{\begin{tabular}{lcc}
\toprule
 & Latent & Gradient\\
System & attn. (\%) & ratio\\
\midrule
\ANAttentionRows
\bottomrule
\end{tabular}}
\endgroup}
\end{table}
\paragraph{Problem-specific reasoning.}
We decode each system's latents with DeepSeek-OCR2 to examine the reasoning content they carry (Table~\ref{tab:analysis-reasoning}). CALR recovers 74\% of the current problem's intermediate values, compared with 7\% when the same decoded text is matched against another problem. Its full-chain character similarity reaches 0.34, compared with 0.07 for RoT and 0.09 for our DLR-SFT reproduction, while answer accuracy reaches 63.5\%. DLR-SFT likewise carries problem-specific intermediate values (0.53 versus 0.13), whereas RoT's recovery is 0.01 in both conditions. This contrast connects the decoded reasoning content to C2 while preserving the distinction from C1: DLR-SFT represents intermediate information that its answer model largely bypasses; CALR both represents and uses it.

Table~\ref{tab:analysis-reader} examines the fidelity of the anchored representation through the frozen external reader. Raw optical features support 98.9\% accuracy, and resampled reference latents retain 96.7\%, a difference of 2.2 points. Latents from the unified-trained producer yield 62.5\%, compared with 0.5\% for a blank page. The reference results establish that the compressed latent representation preserves the answer information needed by the frozen reader. The gap between reference and generated latents thus locates the principal performance bottleneck in problem-conditioned latent generation.

\begin{table}[!htbp]
\centering
\ANpair{.52\linewidth}{.45\linewidth}
{\caption{Decoded reasoning: character similarity, same/other-problem arithmetic-step recovery, and answer accuracy (\%). $^{*}$ denotes our DLR-SFT reproduction.}\label{tab:analysis-reasoning}}
{\begingroup\ANformat{4pt}
\ANfit{\begin{tabular}{lcccc}
\toprule
 & Chain & \multicolumn{2}{c}{Value recovery} & Answer\\
\cmidrule(lr){3-4}
System & similarity & Same & Other & acc.\\
\midrule
\ANReasoningRows
\bottomrule
\end{tabular}}
\endgroup}
{\caption{Frozen external-reader accuracy (\%) across input representations.}\label{tab:analysis-reader}}
{\begingroup\ANformat{7pt}
\newcommand{\ANRaw}{Raw}
\newcommand{\ANReference}{Resampled}
\newcommand{\ANGenerated}{\textbf{Generated}}
\newcommand{\ANBlank}{Blank}
\ANfit{\begin{tabular}{lcc}
\toprule
Source & State & Acc.\\
\midrule
\ANReaderRows
\bottomrule
\end{tabular}}
\endgroup}
\end{table}
\FloatBarrier
\begin{table}[!htbp]
\centering
\ANpair{.52\linewidth}{.45\linewidth}
{\caption{Feedback interventions during generation. Zero and Swap denote zeroing and cross-problem replacement; $\Lambda^{\mathrm{ar}}$ is the accuracy drop.}\label{tab:analysis-feedback}}
{\begingroup\ANformat{1.5pt}
\newcommand{\ANAll}{All 32}
\newcommand{\ANLast}{Last 8}
\newcommand{\ANParallel}{Parallel}
\ANfit{\begin{tabular}{llccccc}
\toprule
 & & \multicolumn{3}{c}{Accuracy (\%)} & \multicolumn{2}{c}{Effect (pp)}\\
\cmidrule(lr){3-5}\cmidrule(l){6-7}
Producer & View & Base & Zero & Swap & $\Lambda_{\mathrm{zero}}^{\mathrm{ar}}$ & $\Lambda_{\mathrm{swap}}^{\mathrm{ar}}$\\
\midrule
\ANFeedbackRows
\bottomrule
\end{tabular}}
\endgroup}
{\caption{Output ($H$) and feedback ($P$) interface ablations. Fixed/tied denotes CALR; MLP is trainable.}\label{tab:analysis-interface}}
{\begingroup\ANformat{1.4pt}
\newcommand{\ANFixed}{Fixed}
\newcommand{\ANFixedTied}{Tied}
\newcommand{\ANMLP}{MLP}
\newcommand{\ANFrozenCopy}{Fixed copy}
\newcommand{\ANTrainable}{Learned}
\ANfit{\begin{tabular}{llccc}
\toprule
 & & & \multicolumn{2}{c}{Effect (pp)}\\
\cmidrule(l){4-5}
$H$ & $P$ & Acc. & $\Lambda_{\mathrm{zero}}^{\mathrm{ar}}$ & $\Lambda_{\mathrm{swap}}^{\mathrm{ar}}$\\
\midrule
\ANInterfaceRows
\bottomrule
\end{tabular}}
\endgroup}
\end{table}
\FloatBarrier
\paragraph{Autoregressive dependence.}
Table~\ref{tab:analysis-feedback} tests whether preceding latent blocks shape subsequent generation. Zeroing or replacing their feedback reduces CALR's accuracy by 25.1 or 18.4 points. The parallel-trained producer, evaluated under the same four-block rollout, remains near its 51.1\% baseline under both interventions. Autoregressive training thus develops a dependence on preceding latent content. This dependence persists when the reader receives only the final block: accuracy falls from 28.2\% to 6.1\% or 9.8\%. Earlier blocks therefore influence what the final block represents, beyond their direct contribution to the reader. Together with the decoded intermediate values, these results show that CALR organizes problem-specific reasoning into a dependent latent sequence.

\paragraph{The role of a fixed interface.}
Table~\ref{tab:analysis-interface} compares output and feedback mappings. The fixed, tied scaled-isometric interface achieves 63.8\% accuracy with zero/swap effects of 25.1/18.4 points. Training the output map reduces accuracy to 59.2\% and the effects to 15.2/13.3; training both maps further reduces them to 56.33\% and 6.2/3.5. A stable interface therefore supports both answer accuracy and prefix dependence. Together, these analyses show that CALR couples answer reliance with problem-specific reasoning, which autoregressive refinement develops through successive latent blocks.

\FloatBarrier
\endgroup

\clearpage
\subsection{Intermediate-State Substitution}
\label{app:intermediate_state_substitution}

We test whether information carried by an intermediate latent block is used in subsequent generation by replacing it with a block generated for another problem. We pair the original problem $x$ with a donor $x'$ that has the same final answer, $y'=y$, but a different first intermediate quantity, $u'_1\neq u_1$. Let $F_x$ denote the composition of the original problem's remaining operations. We select pairs for which applying these operations to the donor's intermediate quantity yields an expected answer distinct from their shared final answer:
\begin{equation}
y_{\mathrm{exp}}=F_x(u'_1),\qquad y_{\mathrm{exp}}\neq y=y'.
\label{eq:calr_expected_answer}
\end{equation}
Producing $y_{\mathrm{exp}}$ therefore requires a result distinct from either problem's original final answer.

With the original question visible to the producer, we replace the second latent block in its feedback with the donor's corresponding block and regenerate the third and fourth blocks. The frozen question-blind external reader receives the original first two blocks followed by the regenerated suffix; the reader-only control instead replaces the second block directly at readout, retaining all other original blocks without regeneration. A further control uses donors with the same intermediate quantity but a different final answer, while self-replacement uses the original block in place of a donor block.

\begin{table}[!htbp]
\centering
\caption{Intermediate-state substitution on two disjoint sets of 120 problems each. Expected denotes the percentage of outputs matching $y_{\mathrm{exp}}$; Orig.\ denotes accuracy against the original answer $y$. Regeneration retains the original first two blocks at readout. All values are percentages.}
\label{tab:calr_state_substitution}
\begingroup
\small
\setlength{\tabcolsep}{5pt}
\renewcommand{\arraystretch}{1.12}
\begin{tabular*}{\linewidth}{@{\extracolsep{\fill}}llcc@{}}
\toprule
Condition & Metric & Set 1 & Set 2\\
\midrule
Feedback substitution + suffix regeneration & Expected & 75.0 & 80.0\\
Reader-input substitution only & Expected & 0.8 & 0.8\\
Same intermediate, different answer & Orig. & 87.5 & 82.5\\
\bottomrule
\end{tabular*}
\endgroup
\end{table}

\paragraph{Results.}
Table~\ref{tab:calr_state_substitution} shows that feedback substitution followed by suffix regeneration produces the expected answer in 75.0\% and 80.0\% of cases across the two sets, respectively, with the original first two blocks retained at readout. Replacing the second block only at the reader input yields the expected answer in 0.8\% of cases on each set. Donors with the same intermediate quantity but different final answers preserve original-answer accuracy at 87.5\% and 82.5\%, matching the self-replacement controls. These results show that substituted intermediate information shapes subsequent latent generation, with its effect reaching the reader through the regenerated suffix.
\FloatBarrier

\end{document}